\documentclass{article}

\usepackage[preprint]{neurips_2026}

\usepackage[utf8]{inputenc} 
\usepackage[T1]{fontenc}    
\usepackage{hyperref}       
\usepackage{url}            
\usepackage{booktabs}       
\usepackage{amsfonts}       
\usepackage{nicefrac}       
\usepackage{microtype}      
\usepackage{xcolor}         

\usepackage{subcaption}
\usepackage{graphicx}
\usepackage{amsmath}
\usepackage{wrapfig}
\usepackage{subcaption}     
\usepackage[most]{tcolorbox} 
\usepackage{listings}
\usepackage{algorithm}      
\usepackage{algpseudocode}  
\title{Prefix Teach, Suffix Fade: Local Teachability Collapse in Strong-to-Weak On-Policy Distillation}

\author{
Kaiyuan Liu\textsuperscript{1,2}, 
    Ziyuan Zhuang\textsuperscript{2}, 
    Yang Bai\textsuperscript{2}, 
    Bing Wang\textsuperscript{3}, 
    Rongxiang Weng\textsuperscript{2}$^{\dagger}$, 
    Jieping Ye\textsuperscript{1} \\
    \textsuperscript{1}{College of Computer Science and Technology, Zhejiang University}, 
    \textsuperscript{2}{Meituan LongCat Team, China} \\
    \textsuperscript{3}{College of Computer Science and Technology, Jilin University} \\ 
    12421281@zju.edu.cn
}

\begin{document}

\maketitle

\begin{abstract}
On-policy distillation (OPD) trains a student model on its own rollouts using dense feedback from a stronger teacher. Prior literature suggests that, provided teacher feedback is available, supervising the full sequence of response tokens should monotonically improve performance. However, we demonstrate that this assumption sometimes fails to hold in strong-to-weak OPD settings. While later segments of a generated trajectory may still exhibit a non-zero teacher-student advantage, they frequently lack the local contrast that makes dense feedback effective for prioritizing student learning. We term this failure mode local teachability collapse.
The resulting principle is straightforward: supervision should concentrate on trajectory regions where the teacher's feedback remains discriminative, rather than uniformly covering the entire response. We operationalize this principle through a trajectory-specific release rule. This rule measures the teacher’s margin over the student’s top-$K$ candidate set, aggregates this margin across NLTK-tokenized sentence segments, and truncates dense OPD supervision upon detecting a BIC-style downward change point.
Experimental results across strong-to-weak distillation tasks using the Qwen3 model family indicate that this release rule consistently outperforms standard full-trajectory OPD across five in-domain benchmarks at various student scales. Furthermore, compared to baseline distillation methods, our approach better preserves model capabilities on out-of-domain task. These results suggest that effective strong-to-weak OPD requires evaluating not only the availability of teacher guidance but also its local utility, ensuring that the generated feedback remains teachable.
\let\thefootnote\relax
\footnotetext{$^{\dagger}$ Corresponding author}
\end{abstract}

\section{Introduction}

On-policy distillation (OPD) trains a student model on its own generated trajectories using feedback from a stronger teacher~\cite{opd}. This on-policy setting is advantageous because the teacher evaluates states the student actually visits, thereby mitigating the exposure bias associated with fixed offline teacher demonstrations. Recent work has improved OPD through reward extrapolation, entropy-aware objectives, support matching, and computationally efficient prefix-style supervision~\citep{exopd,entropy_aware_opd,revisiting_opd,fastopd}. However, a common assumption underlying dense OPD is that once teacher feedback is available for a rollout, applying supervision across the entire generated sequence is uniformly beneficial.

However, we contend that this assumption is overly coarse. While foundation model research typically emphasizes OPD between models of comparable scale, this paper focuses on \textit{strong-to-weak OPD}. This paradigm is increasingly prevalent in on-device modeling and local deployments, where a highly capable teacher supervises a significantly smaller student. In this regime, the critical challenge is not merely whether a preference gap persists between the teacher and the student; rather, the teacher's feedback must remain discriminative over the local region of the rollout being trained. This mirrors a fundamental principle in outcome-driven reinforcement learning (RL): updates are most effective when feedback clearly distinguishes between alternatives, such as the common heuristic of training on queries with a 50\% rollout-accuracy~\citep{rl_acc05_1}. Conversely, samples that are uniformly correct, uniformly incorrect, or otherwise low-contrast provide negligible learning signals~\citep{rl_acc05_1,rl_acc05_2,rl_acc05_3,rl_acc05_dapo}. Consequently, if strong-to-weak OPD applies dense supervision to segments where the teacher's feedback has become nearly flat, it risks optimizing over low-informative noise without providing meaningful local prioritization for the student.

We term this failure mode \textit{local teachability collapse}: a sustained decrease in the teacher's local contrast over the course of a student rollout. In this paper, we first diagnose this phenomenon within a standard strong-to-weak distillation framework from prior work \citep{exopd}, where the capability gap is sufficiently large to expose the effect clearly. Although the mean teacher-student advantage remains nonzero toward the end of a response, the discriminative power of the sampled token advantages decays. This demonstrates that the teacher's global guidance can persist even as dense feedback becomes less informative for identifying which specific actions are relatively better or worse.

To mitigate this issue, we evaluate a simple trajectory-specific release rule. At each prefix, we construct the student's top-$K$ candidate set to represent its reachable space and compute the teacher's margin between the top-1 and top-2 candidates within that set. This nearest-competitor margin serves as a proxy for whether the teacher still provides a sharp local preference signal. We then aggregate this margin over NLTK-tokenized sentence segments~\citep{nltk} and apply a Bayesian Information Criterion (BIC)-style downward change-point rule. Before the detected change point, training relies on the standard OPD advantage objective. After this point, dense OPD supervision is masked out. This experiment isolates the effect of modifying the supervision region while keeping the underlying OPD objective intact.

Across our strong-to-weak distillation experiments, this release rule improves the average score across five in-domain benchmarks from 36.8\% with full OPD to 40.1\% for a 1.7B student. A random-release control yields weaker performance, supporting the need to couple each trajectory with its own local-teachability signal. Furthermore, the same release rule improves aggregate performance over full OPD and other alternative baselines for 4B and 8B students. Finally, the out-of-domain benchmark check suggests that this method better preserves out-of-domain capabilities compared to alternative baselines.

The contribution of this paper is threefold:

\begin{enumerate}
    \item \textbf{Identify local teachability collapse in strong-to-weak OPD}: Motivated by the observation that outcome-RL updates are most useful when feedback distinguishes between alternatives, we analyze whether dense strong-to-weak OPD advantages remain locally discriminative along a student rollout. This perspective reveals \textit{local teachability collapse}, where teacher-student disagreement remains nonzero even as sampled-token advantage dispersion and teacher local contrast decay.
    \item \textbf{A trajectory-specific release rule}: We introduce a mechanism to determine if teacher feedback remains locally actionable for student trajectory. This rule measures the teacher's top-1 and top-2 margin over the student’s reachable candidates, aggregates the signal over sentence segments, and terminates dense supervision after a statistical change point.
    \item \textbf{Empirical validation of the teachability signal}: Across various Qwen3 strong-to-weak distillation settings, our release rule consistently outperforms full OPD on in-domain benchmarks for multiple student sizes and better preserves performance on the out-of-domain benchmark. These results support the central claim of this paper: dense strong-to-weak OPD should consider not only whether a teacher-student advantage exists, but also whether that advantage remains locally teachable.
\end{enumerate}

\section{Related Work}

\paragraph{Off-policy distillation.}
Off-policy distillation addresses distribution mismatch at the trajectory level by modifying the supervision format or data selection process. Standard supervised fine-tuning (SFT) on teacher-generated chain-of-thought traces exhibits high sensitivity to data provenance, distribution, and structural properties. Specifically, performance gains strongly correlate with actions originating directly from the teacher~\cite{sft_d_where_did}. Moreover, SFT fundamentally fails to capture sequence-level distributions and consequently suffers from exposure bias~\cite{sft_d_dasd}, while log-probability-based data selection is frequently confounded by response length~\cite{sft_d_wangbing}.

To mitigate these underlying issues, recent approaches introduce novel alignment strategies. For example, Lion generates increasingly challenging instructions through adversarial loops~\cite{blackbox_opd2}. ORPO-Distill casts cross-architecture distillation as a preference optimization problem~\cite{blackbox_opd4}, whereas DAIL transforms expert solutions into trajectories that are more amenable to student learning~\cite{blackbox_opd5}. Although these methods effectively resolve trajectory-level mismatches, securing reliable token-level supervision from the teacher remains an open challenge. This limitation inherently motivates the transition to on-policy distillation in settings where teacher logits are accessible.

\paragraph{On-policy distillation.}
On-policy distillation (OPD) trains on student-generated trajectories to mitigate exposure bias. Foundational OPD-style frameworks query teacher feedback on student rollouts~\cite{opd}. Subsequent work modifies the OPD objective: G-OPD formulates it as KL-constrained RL, ExOPD uses reward extrapolation~\cite{exopd}, and entropy-aware OPD modulates divergence penalties~\cite{entropy_aware_opd}. Recent analyses identify failure modes such as sparse signals and tokenizer artifacts, motivating top-$K$ support matching~\cite{revisiting_opd}. FastOPD applies OPD only to prefixes for efficiency~\cite{fastopd}, choosing a prefix length or schedule. 
Recent studies increasingly recognize that dense teacher supervision in OPD is not always reliable. For instance, Uni-OPD \cite{uni_opd} observes that token-level teacher guidance can contradict overall trajectory correctness, and proposes an outcome-guided margin calibration to globally restore sequence-level consistency. Distinct from this outcome-centric perspective, our work uncovers a complementary temporal failure mode: local teachability collapse. We demonstrate that the local discriminative utility of dense supervision inherently degrades as the rollout progresses in strong-to-weak settings.

OPD has recently expanded to multimodal alignment (vision-language reasoning~\cite{mllm_opd1}, video grounding~\cite{mllm_opd2}, and speech LLMs~\cite{mllm_opd3}). Additionally, self-distillation methods construct on-policy supervision without external teachers via self-play, privileged-context tutoring, or historical checkpoints~\cite{self_opd1,self_opd2,self_opd3,self_opd4,self_opd11,self_opd12}. While these directions broaden how teacher signals are obtained, our focus is complementary. Motivated by settings such as on-device modeling and local deployment, we investigate a specific scenario: when white-box logits from a stronger external teacher are available, does this dense teacher signal remain locally discriminative enough to effectively supervise a weak student's rollout.

\section{Local Teachability Analysis}
\label{sec:analysis_preliminaries}

\paragraph{Preliminaries.}
In on-policy distillation (OPD), the student samples its own response \(y=(y_1,\ldots,y_T)\) for an input \(x\), and the teacher is evaluated on the same visited prefixes \((x,y_{<t})\)~\citep{opd,exopd,mimo_v2,think_machine_lab}. This makes the supervision on-policy: the teacher does not provide a separate offline trajectory, but instead scores the states actually induced by the student.

There are two primary approaches to formulating the loss in On-Policy Distillation: minimizing the Kullback-Leibler (KL) divergence over a full or partial vocabulary, or optimizing a reinforcement learning (RL) objective via advantage estimation. In this work, we adopt the latter formulation, as it is more memory-efficient and widely utilized in large-scale industrial deployments, such as Alibaba Qwen3~\citep{qwen3_report}, Tencent HunYuan~\citep{exopd}, XiaoMi MiMo-v2~\citep{mimo_v2}, ZhiPu GLM-5~\citep{glm5}, NVIDIA~\citep{nvidia} and ThinkMachineLab~\citep{think_machine_lab}. Under this framework, the standard OPD signal is defined as the log-probability difference between the teacher and the student for a sampled token. For each sampled response token \(y_t\), the teacher-student advantage is
\[
A_t = \log p_T(y_t \mid x, y_{<t}) - \log p_S(y_t \mid x, y_{<t}).
\]
The signed mean of \(A_t\) measures teacher-student disagreement on the sampled trajectory. A positive $A_t$ indicates that the teacher assigns a higher probability to the action than the student, suggesting the student should increase its likelihood, while a negative value implies the opposite. The training loss incorporates $A_t$ as dense feedback along the sampled trajectory, typically by weighting the student's log-probabilities within a policy-gradient objective.

\paragraph{Diagnosing local teachability collapse.}
\label{sec:analysis}

\begin{figure}[t]
    \centering
    \begin{subfigure}{0.45\linewidth}
        \centering
        \includegraphics[width=\linewidth]{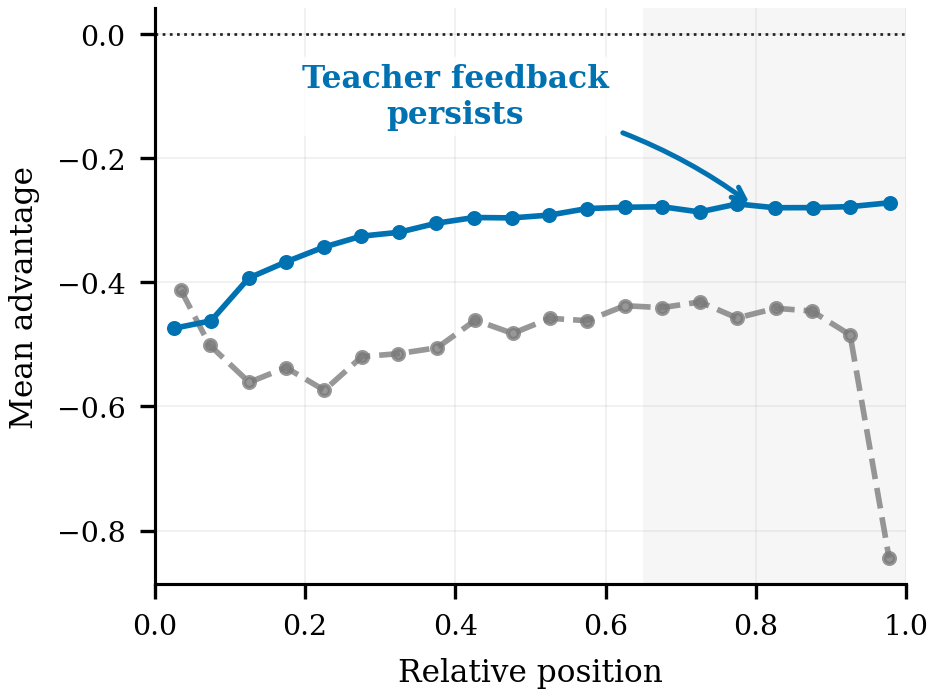}
        \caption{Mean advantage.}
        \label{fig:local_teachability_global}
    \end{subfigure}
    \hfill
    \begin{subfigure}{0.45\linewidth}
        \centering
        \includegraphics[width=\linewidth]{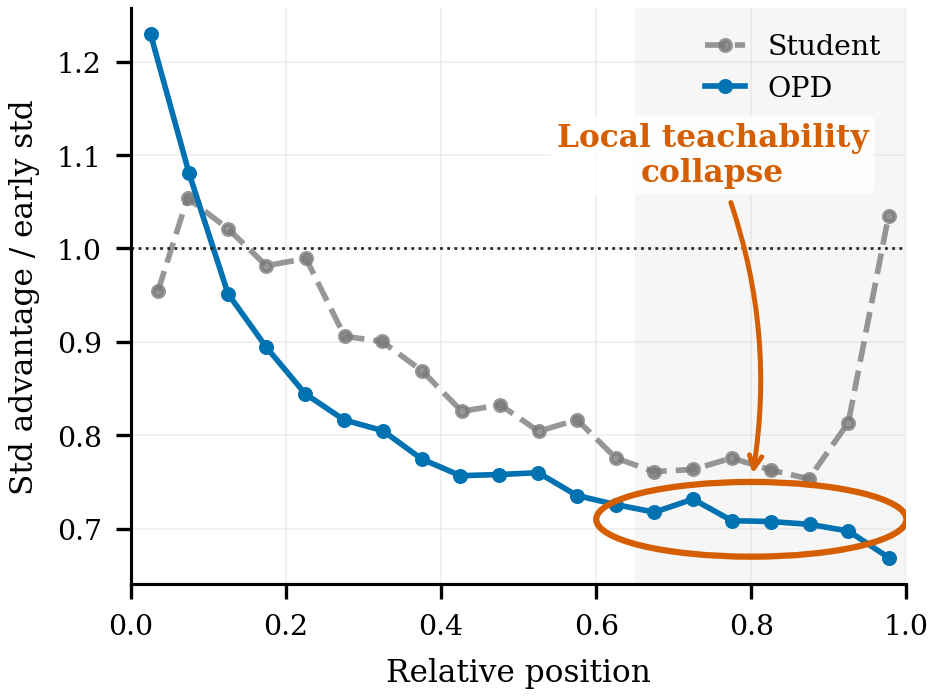}
        \caption{Advantage dispersion.}
        \label{fig:local_teachability_contrast}
    \end{subfigure}

    \caption{\textbf{The presence of a guidance signal does not guarantee high-value local teachability.} (a) The mean teacher-student advantage remains nonzero across late response regions, indicating that the teacher has not simply ceased guiding the student. (b) The within-bin standard deviation of $A_t$, normalized by its early-region value, steadily decays throughout the rollout. This decrease in sampled-path dispersion is a diagnostic indicator that the feedback becomes less token-specific, yielding relatively lower instructional value compared to the earlier dense feedback.}
    \label{fig:local_teachability_combined}
\end{figure}

OPD assumes that teacher disagreement remains uniformly informative throughout the sampled response. In strong-to-weak OPD, this assumption can break down because a preference gap may persist even as the teacher provides diminishing contrast across later segments. This issue is significant for the same reason that outcome-driven RL benefits most from non-degenerate feedback groups: when feedback fails to distinguish between alternative or partial trajectories, it yields a weaker learning signal~\citep{rl_acc05_1,rl_acc05_2,rl_acc05_3,rl_acc05_dapo}. We term the OPD variant of this failure mode \emph{local teachability collapse}: a sustained decline in the teacher's local discriminability along a rollout.

We initially examine this phenomenon using the strong-to-weak distillation framework established in previous OPD study~\citep{exopd}. The purpose of this setting is not to demonstrate uniform scaling curves across all student sizes; rather, the pronounced performance gap between the teacher (Qwen3-30B-A3B-Instruct-2507) and student (Qwen3-1.7B) serves to expose the local-support mismatch. Section~\ref{sec:exp_scaling} then investigates whether our release rule formulated from this analysis generalizes to more capable students.

Our analysis compares the base student model (denoted as \textit{Student} in Figure~\ref{fig:local_teachability_combined}) with its OPD-optimized counterpart (denoted as \textit{OPD} in Figure~\ref{fig:local_teachability_combined}). While the base student provides a baseline for the initial on-policy distribution, the OPD model serves as our primary subject for evaluating policy convergence. This comparison allows us to isolate the generic characteristics of weak-model rollouts from the patterns that emerge or persist specifically following OPD optimization.

To quantify the value of local teachability, we sample responses from both models across 500 training instances. For each trajectory, we compute the teacher and student log-probabilities at every decoding step to derive the pointwise advantage, $A_t$. We partition these sequences into 20 discrete segments based on their normalized temporal position within the trajectory. For each segment, we compute two primary bin-wise statistics:

\begin{enumerate}
\item \textbf{Mean Advantage:} This represents the aggregate teacher preference relative to the student's current policy, capturing whether the teacher continues to actively guide the student toward optimization.
\item \textbf{Standard Deviation of Advantage:} This measures the local token-level variance, reflecting the discriminative power of the teacher's guidance signal across adjacent actions.
\end{enumerate}

Figure~\ref{fig:local_teachability_combined} disentangles two questions that are frequently conflated. Panel (a) examines whether the teacher and student still disagree on sampled tokens. The results indicate that they do: for the OPD-optimized model, the late mean advantage remains at approximately $A_t \approx -0.3$. Panel (b) investigates whether the dense feedback retains token-specific variation. This effect is noticeably weaker, as the normalized standard deviation of $A_t$ decreases substantially in later regions. Consequently, the teacher's global guidance can persist even when the token-level signal becomes less informative for distinguishing the relative quality of local actions. A representative log-probability trace in Figure~\ref{fig:appendix_local_teachability_case} exhibits the same qualitative pattern. This initial diagnostic serves as the primary motivation for the rest of our work. By examining the sampled advantage sequence, we illustrate that dense supervision applied to the entire response often yields diminishing local information.

\begin{wrapfigure}[20]{r}{0.50\columnwidth}
    \centering
    \includegraphics[width=0.49\columnwidth]{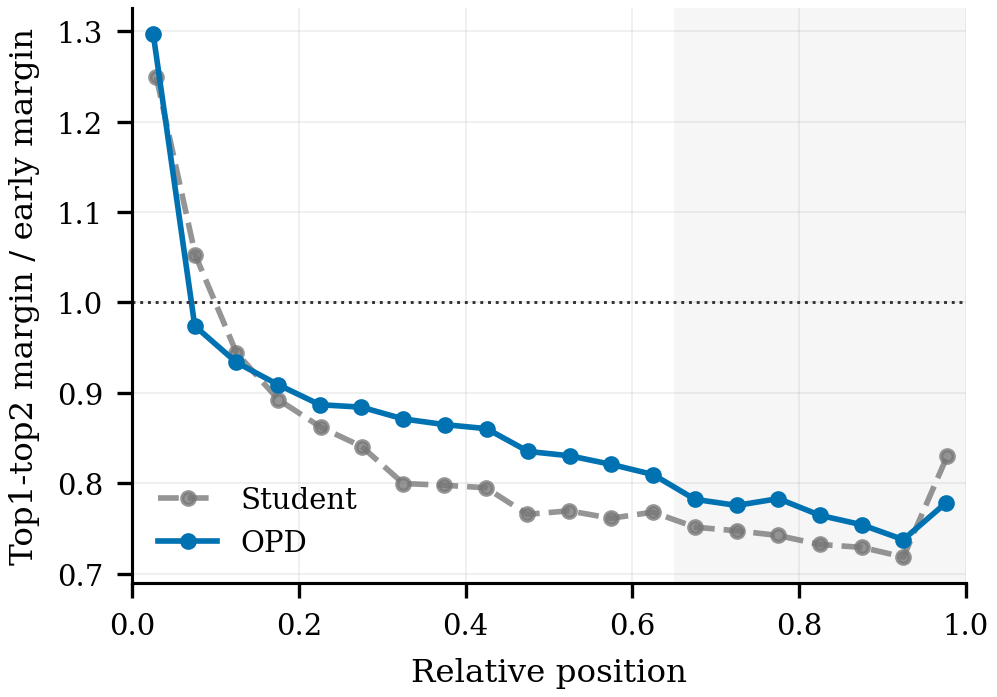}
    \caption{\textbf{Teacher margin over student top-$K$ candidates.} The normalized margin declines along the response, providing the local signal that is later aggregated over sentence segments for release.}
    \label{fig:method_margin_curve}
\end{wrapfigure}

\paragraph{Trajectory-specific release rule.}
\label{sec:analysis_method}

The preceding diagnostic, combined with the established insight from outcome-driven RL that high-contrast advantages are more useful for policy updates while low-contrast advantages can often be discarded \citep{rl_acc05_1,rl_acc05_2,rl_acc05_3}, suggests a natural strategy: dropping supervision when the chunk-level $\mathrm{std}(A_t)$ becomes low. In practice, however, we observe that the chunk-level $\mathrm{std}(A_t)$ is unstable because it measures variance solely across the specific tokens sampled by the student. This narrow evaluation places undue emphasis on individual sampled trajectories and ignores the underlying capacity of the teacher to guide the student at each prefix state, thereby yielding an inconsistent measure of local teachability. This limitation is particularly pronounced during the early stages of student training. As illustrated in Figure~\ref{fig:local_teachability_contrast}, when a trajectory reaches the final answer summary stage (the last bin), the teacher might abruptly provide discriminative action-level feedback for an obviously incorrect answer, even if local teachability has already collapsed during earlier segments of the solution. Consequently, before determining where to apply supervision, a more robust measure must answer a fundamental question: given the current prefix state, can the teacher still discriminate among the actions the student is likely to take? Answering this requires analyzing the local support of the student policy rather than relying exclusively on the sampled token sequence. We revisit this distinction in Section~\ref{sec:exp_support_size_adv_check}.

We therefore estimate chunk-level teachability using a state-local teacher margin and subsequently aggregate it over NLTK-tokenized sentence segments \citep{nltk}. At prefix $(x,y_{<t})$, let$$\mathcal{C}_t=\operatorname{TopK}_{a}\; p_S(a\mid x,y_{<t})$$be the student's top-$K$ candidate set, which represents the student's reachable action space at that prefix. We rank candidates in $\mathcal{C}_t$ using teacher log-probabilities and define$$a_t^{(1)}=\arg\max_{a\in\mathcal{C}_t}\log p_T(a\mid x,y_{<t}),\qquad
a_t^{(2)}=\arg\max_{a\in\mathcal{C}_t\setminus\{a_t^{(1)}\}}\log p_T(a\mid x,y_{<t}).$$The token-level local contrast proxy is the teacher's nearest-competitor margin$$M_t=\log p_T(a_t^{(1)}\mid x,y_{<t})-\log p_T(a_t^{(2)}\mid x,y_{<t}).$$Unlike $A_t$, this margin does not score the sampled token itself. Instead, it evaluates whether the teacher can still express a sharp pedagogical preference within the student's reachable support. Following sentence-level aggregation, a sustained margin drop indicates that the corresponding segment has diminished point-level teaching value.

The margin curve in Figure~\ref{fig:method_margin_curve} reflects the broader decline implied by advantage dispersion, but it evaluates a distinct quantity: the teacher's ranking over candidates reachable by the student under the same prefix. This distinction is central to our release rule. While the realized-path statistic $\mathrm{std}(A_t)$ identifies the underlying symptom, $M_t$ translates this intuition into a state-local control signal by assessing whether the teacher maintains a clear preference among the student's viable candidates.

Although we fixed the number of bins at 20 during our analysis, the significant variance in trajectory lengths makes it difficult to define an appropriate bin parameter for actual training. To address this, we adopt a more natural approach using NLTK sentence segmentation \citep{nltk}. Let $\mathcal{B}_i$ denote the set of tokens within sentence segment $i$. The segment-level teachability signal is defined as follows:$$S_i=\log\!\left(1+\frac{1}{|\mathcal{B}_i|}\sum_{t\in\mathcal{B}_i}M_t\right).$$The logarithmic function is applied primarily to stabilize the numerical scale. Furthermore, sentence-level aggregation mitigates transient token-level artifacts while aligning the release decision with natural linguistic boundaries.

We then use the sequence \(\{S_i\}_{i=1}^n\) to decide whether the rollout should keep full OPD supervision or be released after a local-teachability drop. This can be cast as a simple two-model comparison. The null hypothesis is that the trajectory has one stable teachability level, while the alternative is that there is a single downward shift after some segment \(\tau\). The no-change model is

\[
H_0:\quad S_i=\mu+\epsilon_i,
\]
and the one-drop model is
\[
H_1(\tau):\quad
S_i=
\begin{cases}
\mu_{\mathrm{pre}}+\epsilon_i, & i\le \tau,\\
\mu_{\mathrm{post}}+\epsilon_i, & i>\tau,
\end{cases}
\qquad \mu_{\mathrm{pre}}>\mu_{\mathrm{post}} .
\]
For each candidate \(\tau\), we compute the residual sum of squares:
\[
\mathrm{RSS}_0=\sum_i(S_i-\bar S)^2,
\]
\[
\mathrm{RSS}_1(\tau)
=
\sum_{i\le\tau}(S_i-\bar S_{\le\tau})^2
+
\sum_{i>\tau}(S_i-\bar S_{>\tau})^2.
\]

We use a profiled RSS-BIC criterion, where the noise variance is absorbed into the residual term rather than counted as an additional explicit parameter:
\[
\operatorname{BIC}_0
=
n\log\left(\frac{\mathrm{RSS}_0+\epsilon}{n}\right)
+
k_0\log n,
\]
\[
\operatorname{BIC}_1(\tau)
=
n\log\left(\frac{\mathrm{RSS}_1(\tau)+\epsilon}{n}\right)
+
k_1\log n.
\]
Under this convention, the no-change model has one free mean parameter, so \(k_0=1\). The one-drop model has two segment means and one discrete change-point parameter, so \(k_1=3\).

We select the best downward change point by minimizing the BIC among candidates satisfying \(\mu_{\mathrm{pre}}>\mu_{\mathrm{post}}\):
\[
\tau^\star
=
\arg\min_{\tau:\,\bar S_{>\tau}<\bar S_{\le\tau}}
\operatorname{BIC}_1(\tau).
\]
A release is accepted when the best one-drop model improves over the no-change model:
\[
\operatorname{BIC}_1(\tau^\star)<\operatorname{BIC}_0.
\]
If no downward candidate improves the no-change model, the trajectory defaults to full OPD supervision.

Given the accepted change point $\tau^\star$, we retain only the tokens preceding the corresponding segment boundary. Let $q_t\in\{0,1\}$ denote this prefix mask, and let $l_t$ denote the original loss mask covering the entire trajectory. To preserve the per-sample loss mass after truncation, we rescale the retained prefix to ensure the overall consistency of the gradient magnitude:
\[
A'_t
=
A_t\cdot q_t\cdot
\frac{\sum_t l_t}
{\sum_t l_t q_t}.
\]
Consequently, the release rule modifies where teacher-guided OPD supervision is applied, rather than altering the total supervision weight assigned to a trajectory. This approach focuses learning on the prefix region where the teacher still provides a clear state-local preference within the student's reachable action space, while attenuating the influence of low-teachability suffixes.

\begin{figure}[t]
\centering
\begin{subfigure}{0.31\linewidth}
\centering
\includegraphics[width=\linewidth]{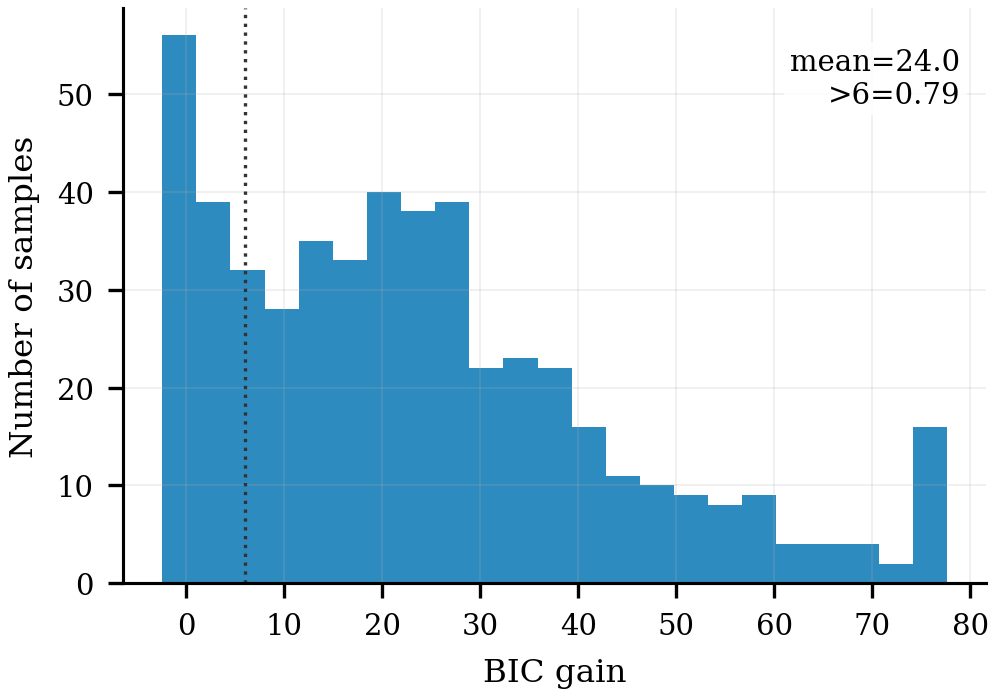}
\caption{BIC gain.}
\label{fig:method_proxy_bic}
\end{subfigure}
\hfill
\begin{subfigure}{0.31\linewidth}
\centering
\includegraphics[width=\linewidth]{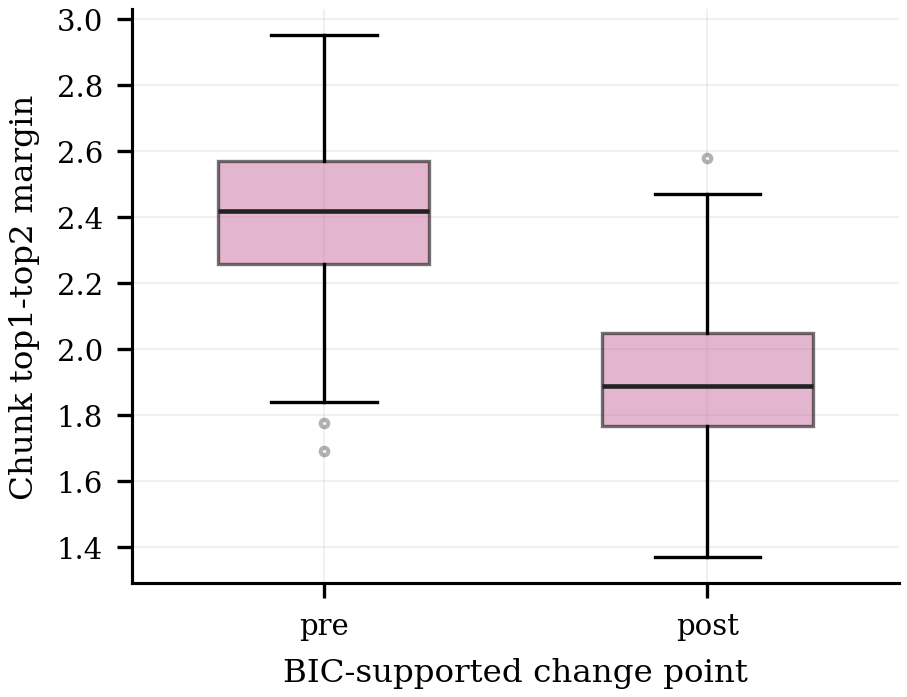}
\caption{Pre/post margin.}
\label{fig:method_pre_post_margin}
\end{subfigure}
\hfill
\begin{subfigure}{0.31\linewidth}
\centering
\includegraphics[width=\linewidth]{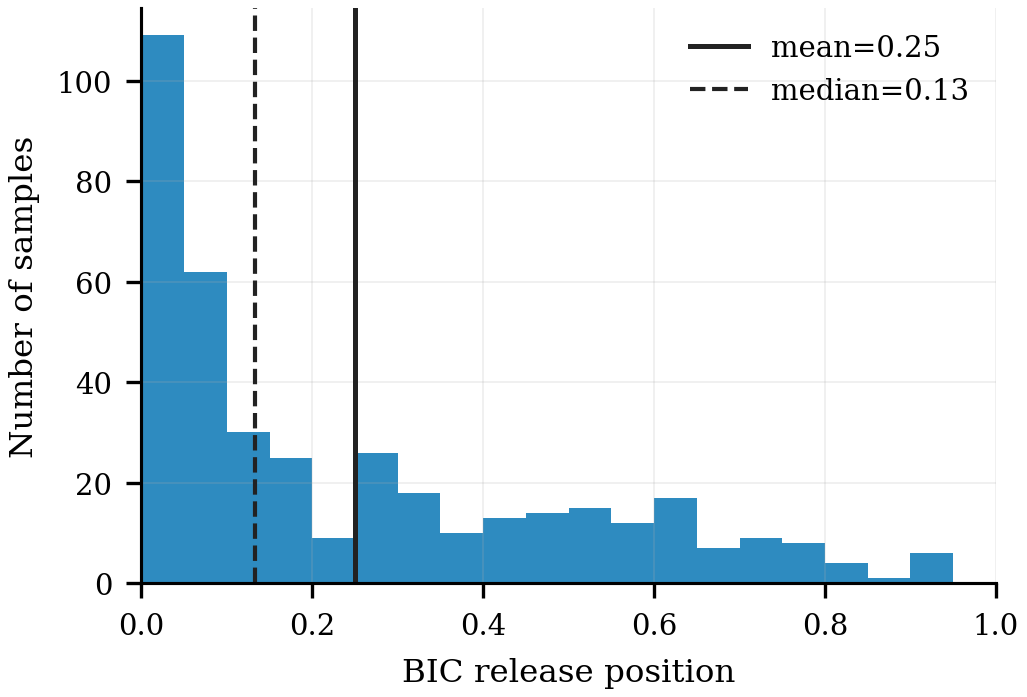}
\caption{Release distribution.}
\label{fig:method_release_distribution}
\end{subfigure}
\caption{\textbf{The release rule detects trajectory-specific drops in local contrast.} Statistics are computed on the exported OPD rollouts using the teacher's top-1 and top-2 margin over the student's top-\(K\) candidates. (a) The profiled RSS-BIC gain, \(\operatorname{BIC}_0-\min_\tau \operatorname{BIC}_1(\tau)\), is positive for most samples. The mean gain is 24.0, and 79\% of the samples accept a significant downward BIC-improving release with a gain greater than 6. (b) For accepted samples, the average chunk-level margin decreases from the pre-release region to the post-release region. (c) Relative release positions vary across trajectories, with a mean of 0.25 and a median of 0.13, demonstrating that the rule is not merely a fixed prefix in disguise.}
\label{fig:method_bic_evidence}
\vspace*{-2\baselineskip}
\end{figure}

Figure~\ref{fig:method_bic_evidence} characterizes the release rule itself. The BIC gain demonstrates when a one-drop summary is preferred; the pre/post comparison confirms that accepted changes are downward; and the release distribution illustrates substantial sample-level variation. Appendix~\ref{sec:appendix_release_algorithm} provides the implementation details, and Appendix~\ref{sec:appendix_theoretical_motivation} gives a signal-to-noise interpretation of why locally low-contrast suffix supervision can reduce the usefulness of dense OPD updates. Figure~\ref{fig:method_release_case} and Appendix~\ref{sec:appendix_release_case_study} present some representative release cases.

The proposed rule operates as a practical heuristic rather than a theoretical guarantee that all post-release tokens are detrimental. Designed to validate the central claim of this paper, its objective is to maintain standard OPD as long as sentence segments retain sufficient teacher contrast to be informative, and to halt dense supervision once the sentence-level margin curve exhibits a substantial downward shift. The following section evaluates whether this trajectory-specific release mechanism effectively improves training performance.
\vspace*{-0.5\baselineskip}
\section{Experiments}
\label{sec:experiments}
\vspace*{-0.5\baselineskip}

The preceding analysis raises a concrete empirical question: if dense supervision is halted when the teacher's local teachability declines, does OPD yield better performance than supervising the entire response or selecting a global prefix length? We address this question in four steps. First, we compare full OPD, fixed-prefix OPD, and the trajectory-specific release rule within the strong-to-weak setting used in prior work~\citep{exopd}. Second, we examine whether the release point must remain coupled to each trajectory. Third, we conduct targeted sanity checks regarding the support size and the $\mathrm{std}(A_t)$ diagnostic. Finally, motivated by strong-to-weak deployment settings, we evaluate whether this rule remains effective for larger student models.

\vspace*{-0.5\baselineskip}
\subsection{Experimental Setup}
\label{sec:exp_setup}
\vspace*{-0.5\baselineskip}

Our main setting follows strong-to-weak OPD~\citep{opd_survey,exopd} with Qwen3-1.7B-Non-Thinking as the student and Qwen3-30B-A3B-Instruct-2507 as the teacher. This setup reflects the regime emphasized in the introduction: a substantially stronger teacher supervises a smaller deployable student. Training prompts are sampled from DeepMath~\citep{trainset_deepmath}. We evaluate accuracy on AIME25~\citep{testset_opencompass_aime2025}, HMMT25-Feb~\citep{testset_hmmt}, HMMT25-Nov~\citep{testset_hmmt}, MATH500~\citep{testset_math500}, and Minerva~\citep{testset_minerva}; \textsc{Avg} is the unweighted average over these five math benchmarks. GPQA~\citep{testset_gpqa_diamond} is reported separately as an out-of-domain sanity indicator and is not included in \textsc{Avg}.

We compare the following methods in Table \ref{tab:main_results_1p7b}. \textbf{Student} represents the original weak model, and \textbf{Teacher} is the strong reference model. \textbf{OPD} applies dense supervision across the entire generated response. \textbf{ExOPD} is a reward-extrapolated OPD baseline \citep{exopd}. It modifies the OPD objective, whereas the release variants detailed below alter only the supervision region. \textbf{Fixed-prefix OPD} maintains the standard OPD objective but halts supervision after a global prefix $k\in\{1024,2048,4096,8192\}$. This approach provides a retrospective sweep across prefix lengths, drawing inspiration from prefix supervision methods such as FastOPD \citep{fastopd}. \textbf{Random release} preserves the batch-level release point distribution of the trajectory-specific rule but randomly permutes these release points across samples. \textbf{Ours} computes the margin between the teacher's top-1 and top-2 predictions over the student's top-$K$ candidates, aggregates this margin across NLTK-tokenized sentence segments, and ceases supervision following a downward profiled RSS-BIC improvement.

It is important to note that our release computation serves primarily to validate the analysis presented in this paper. This diagnostic pass should not be interpreted as an efficiency claim comparable to FastOPD \citep{fastopd}. Instead, it represents an exploration of teacher signals: we decouple these signals into existence and value to investigate whether relatively low-value signals influence the training process. Specifically, our current implementation selects a supervision region only after performing full-rollout diagnostics and does not inherently shorten the generation process. Optimizing training efficiency is therefore beyond the scope of this work. Further empirical validation across additional domains is provided in Appendix~\ref{sec:appendix_code_domain_results}.

\vspace*{-0.5\baselineskip}
\subsection{Main Results: Feedback Importance Varies Across Sequences}
\label{sec:exp_main}
\vspace*{-0.5\baselineskip}

\begin{table*}[t]
\centering
\small
\setlength{\tabcolsep}{5pt}
\caption{Main strong-to-weak OPD results. Bold and underline mark the best and second-best trained methods for each metric.}
\label{tab:main_results_1p7b}
\resizebox{\textwidth}{!}{%
\begin{tabular}{lccccccc}
\toprule
Method & AIME25 & HMMT-Feb & HMMT-Nov & MATH500 & Minerva & \textsc{Avg} & GPQA \\
\midrule
Teacher & 61.2\% & 42.4\% & 57.9\% & 98.0\% & 51.7\% & 62.2\% & 70.4\% \\
Student & 11.4\% & 6.6\% & 4.5\% & 74.0\% & 27.9\% & 24.9\% & 31.1\% \\
\hline
OPD~\citep{opd} & 28.4\% & 16.2\% & 15.9\% & 86.2\% & 37.1\% & 36.8\% & 40.7\% \\
ExOPD~\citep{exopd} & 30.8\% & 16.4\% & 17.4\% & \underline{89.2\%} & \textbf{41.2\%} & 39.0\% & 36.6\% \\
Fixed-prefix OPD-1024~\citep{fastopd} & 30.1\% & 18.9\% & 20.6\% & 87.8\% & 35.9\% & 38.7\% & 40.7\% \\
Fixed-prefix OPD-2048~\citep{fastopd} & 30.6\% & 18.5\% & \underline{20.9\%} & \underline{89.2\%} & 37.3\% & 39.3\% & \underline{41.2\%} \\
Fixed-prefix OPD-4096~\citep{fastopd} & \textbf{31.6\%} & \textbf{19.3\%} & 20.4\% & 86.8\% & \underline{38.8\%} & \underline{39.4\%} & 38.6\% \\
Fixed-prefix OPD-8192~\citep{fastopd} & 30.7\% & 17.4\% & 18.4\% & 88.8\% & 37.3\% & 38.6\% & 38.6\% \\
\hline
Random release & 29.2\% & 18.0\% & 18.9\% & 88.0\% & 36.0\% & 38.0\% & 37.4\% \\
Ours & \underline{31.5\%} & \underline{19.1\%} & \textbf{21.9\%} & \textbf{89.4\%} & \underline{38.8\%} & \textbf{40.1\%} & \textbf{42.4\%} \\
\bottomrule
\end{tabular}
}
\vspace*{-2\baselineskip}
\end{table*}

Table~\ref{tab:main_results_1p7b} demonstrates that on-policy teacher feedback provides significant value, as full OPD improves the average math performance from 24.9\% to 36.8\%. However, the results also suggest that dense supervision across the entire response is not always optimal. Each fixed-prefix configuration outperforms full OPD on the \textsc{Avg} metric, indicating that truncating supervision at later stages can be beneficial. Notably, the optimal prefix length varies by benchmark, and the best performance in these rows relies on a globally selected cutoff found through hyperparameter sweeping.

Our trajectory-specific release rule reaches 40.1\% on \textsc{Avg}, improving over full OPD by 3.3 points and modestly exceeding the best fixed-prefix row without choosing a global token index. GPQA follows the same direction in this setting, suggesting that the release rule better preserves out-of-domain capability than the compared alternatives in these runs. The key point is that the local-teachability signal can reduce reliance on global prefix selection while preserving the standard OPD objective.

\vspace*{-0.5\baselineskip}
\subsection{Trajectory Coupling: More Than Retaining Fewer Tokens}
\label{sec:exp_mechanism}
\vspace*{-0.5\baselineskip}

The random-release control experiment investigates whether the performance gains stem merely from retaining a specific volume of supervision. They do not: randomly permuting release points across samples reduces the Avg score from 40.1\% to 38.0\% and degrades GPQA performance. This result validates the central mechanism, suggesting that the release point must remain coupled to the trajectory’s specific local-teachability curve. This coupling effect is also visible in the rollout diagnostics introduced in Section~\ref{sec:analysis_method}. There, Figure~\ref{fig:method_bic_evidence} shows that the teacher-margin signal yields sample-specific downward changes rather than a single shared cutoff. The empirical gain is tied to preserving the trajectory-specific local-teachability signal, not merely to choosing a shorter average prefix.

Fixed-prefix OPD~\citep{fastopd} serves as a critical baseline for our work. While it demonstrates that full dense OPD is excessively coarse, it treats the supervision boundary as a global length prior. In contrast, our trajectory-specific rule conditions the boundary on the teacher-margin curve of each individual rollout. Figure~\ref{fig:prefix_release_token_vs_avg} provides a descriptive validation of this approach: fixed-prefix model checkpoints with rollouts that imply later diagnostic release positions generally demonstrate stronger downstream performance. Although this observation is based on only four prefix checkpoints, it serves as a mechanical illustration of the underlying logic rather than a standalone statistical claim.

\vspace*{-0.5\baselineskip}
\subsection{Sanity Checks: Support Size and Advantage Dispersion}
\label{sec:exp_support_size_adv_check}
\vspace*{-0.5\baselineskip}

For the release-rule experiments, the teacher signal is defined as the margin between the top-1 and top-2 nearest competitors. Table~\ref{tab:support_size_adv_check} maintains this margin definition while investigating two specific questions under the NLTK sentence-segmented profiled-RSS-BIC procedure. First, we examine whether the student's support set size $K$ introduces instability into the results. Second, we evaluate whether the realized-path diagnostic $\text{std}(A_t)$ can effectively substitute for the state-local margin as the release signal.

\begin{table*}[t]
\centering
\scriptsize
\setlength{\tabcolsep}{10pt}
\caption{Support-size and advantage-dispersion checks for our release rule. All support-set rows use the fixed teacher top-1/top-2 margin; $\text{std}(A_t)$ applies the same profiled-RSS-BIC procedure.}
\label{tab:support_size_adv_check}
\begin{tabular}{lccccccc}
\toprule
Variant & AIME25 & HMMT-Feb & HMMT-Nov & MATH500 & Minerva & \textsc{Avg} & GPQA \\
\midrule
OPD~\citep{opd} & 28.4\% & 16.2\% & 15.9\% & 86.2\% & 37.1\% & 36.8\% & 40.7\% \\
\hline
\(\mathrm{std}(A_t)\) release & 30.3\% & 17.7\% & 19.4\% & 88.4\% & 38.1\% & 38.8\% & 40.9\% \\
\(K=2\) support & 30.1\% & \underline{18.8\%} & 20.8\% & 87.2\% & 36.8\% & 38.7\% & \underline{42.2}\% \\
\(K=4\) support & \underline{31.5\%} & \textbf{19.1\%} & \textbf{21.9\%} & \textbf{89.4\%} & \textbf{38.8\%} & \textbf{40.1\%} & \textbf{42.4\%} \\
\(K=16\) support & \textbf{32.3\%} & 18.0\% & 19.4\% & \underline{88.8\%} & \underline{38.2\%} & \underline{39.3\%} & 41.6\% \\
\(K=32\) support & 30.4\% & \underline{18.8\%} & \underline{20.9\%} & 87.6\% & 37.5\% & 39.0\% & 41.9\% \\
\bottomrule
\end{tabular}
\vspace*{-1\baselineskip}
\end{table*}

The support-size rows are close in aggregate, suggesting that the result is not driven by one brittle support choice. We use \(K=4\) for the other results because it gives the strongest \textsc{Avg} in this check while keeping the local action set focused on the student's most competitive next-token candidates. The \(\mathrm{std}(A_t)\) release also improves over full OPD, which is consistent with advantage dispersion being a useful symptom. It remains below the top1--top2 margin with \(K=4\), supporting the distinction from Section~\ref{sec:analysis_method}: advantage dispersion diagnoses the problem on a realized path, while the teacher margin asks a state-local ranking question. Figure~\ref{fig:topk_support_margin_sensitivity} further shows that the normalized teacher-margin curves remain similar across support sizes.

\vspace*{-0.5\baselineskip}
\subsection{Scaling the Release Rule to Stronger Students}
\label{sec:exp_scaling}
\vspace*{-0.5\baselineskip}

Finally, we test whether the same rule remains useful as the strong-to-weak gap narrows. We repeat the protocol with Qwen3-4B and Qwen3-8B students under the same Qwen3-30B-A3B-Instruct-2507 teacher and DeepMath training queries. For all FastOPD experiments, we set the prefix length to 2048, as this configuration yielded the best-balanced in-domain and out-of-domain performance in the main results. Table~\ref{tab:scaling_results} shows that the average-level pattern persists with larger students.

\begin{table*}[t]
\centering
\small
\setlength{\tabcolsep}{5pt}
\caption{Scaling results with stronger students. The same release rule is applied across all scales. \textsc{Avg} denotes the mean performance across five math benchmarks, while GPQA represents out-of-domain evaluation. Bold and underlined values indicate the best and second-best trained methods for each metric within each student scale, respectively.}
\label{tab:scaling_results}
\resizebox{\textwidth}{!}{%
\begin{tabular}{lccccccc}
\toprule
Method & AIME25 & HMMT-Feb & HMMT-Nov & MATH500 & Minerva & \textsc{Avg} & GPQA \\
\midrule
Teacher & 61.2\% & 42.4\% & 57.9\% & 98.0\% & 51.7\% & 62.2\% & 70.4\% \\
\hline
\multicolumn{8}{c}{\textit{\textbf{Student: Qwen3-4B}}} \\
Student & 19.4\% & 11.6\% & 9.2\% & 83.6\% & 39.0\% & 32.5\% & 48.7\% \\
OPD~\citep{opd} & 49.9\% & 30.7\% & 38.4\% & \textbf{94.4\%} & 44.7\% & 51.6\% & \underline{55.6\%} \\
ExOPD~\citep{exopd} & 51.8\% & 30.0\% & 40.3\% & 93.4\% & \textbf{45.6\%} & 52.2\% & 43.9\% \\
FastOPD~\citep{fastopd} & \underline{52.3\%} & \underline{32.4\%} & \textbf{41.5\%} & 93.0\% & 43.4\% & \underline{52.5\%} & 55.1\% \\
Ours & \textbf{54.7\%} & \textbf{34.9\%} & \underline{40.7\%} & \underline{94.0\%} & \underline{45.2\%} & \textbf{53.9\%} & \textbf{57.3\%} \\
\midrule
\multicolumn{8}{c}{\textit{\textbf{Student: Qwen3-8B}}} \\
Student & 21.6\% & 11.8\% & 9.2\% & 84.6\% & 42.1\% & 33.8\% & 40.2\% \\
OPD~\citep{opd} & 53.3\% & 29.3\% & 42.2\% & \textbf{95.4\%} & 45.2\% & 53.1\% & 56.8\% \\
ExOPD~\citep{exopd} & 53.2\% & 29.1\% & \textbf{45.8\%} & \underline{94.0\%} & \textbf{46.3\%} & 53.7\% & 39.1\% \\
FastOPD~\citep{fastopd} & \textbf{55.3\%} & \textbf{34.5\%} & 44.7\% & \underline{94.0\%} & 45.0\% & \underline{54.7\%} & \underline{57.8\%} \\
Ours & \underline{55.1\%} & \underline{33.4\%} & \underline{45.7\%} & \textbf{95.4\%} & \underline{46.1\%} & \textbf{55.2\%} & \textbf{58.6\%} \\
\bottomrule
\end{tabular}
}
\vspace*{-2\baselineskip}
\end{table*}

\vspace*{-0.5\baselineskip}
\section{Conclusion}
\label{sec:conclusion}
\vspace*{-0.5\baselineskip}

This paper investigates strong-to-weak on-policy distillation, demonstrating that dense supervision should fade when the teacher’s feedback loses its local teachability. 
Specifically, we identify \emph{local teachability collapse}, demonstrating that a nonzero teacher-student advantage can coexist with declining local contrast. To address this, we evaluate a trajectory-specific release rule that halts dense on-policy distillation (OPD) supervision following a BIC-style drop in the teacher's margin over the student's reachable candidates. Across multiple strong-to-weak distillation settings, this rule consistently outperforms standard full-trajectory OPD. These results validate the central message of this paper: effective dense OPD must consider not only whether a teacher-student advantage exists, but also whether that advantage remains locally actionable for the student.

\bibliographystyle{plain}
\bibliography{paper}


\clearpage

\appendix

\section{Technical Appendices and Supplementary Material}

\subsection{Release Rule Details}
\label{sec:appendix_release_algorithm}

The algorithm below summarizes the release rule used in the experiments. It is deliberately small: it uses the student's local top-\(K\) support, a nearest-competitor teacher margin, NLTK sentence-level aggregation, a profiled RSS-BIC one-drop test, and retained-prefix loss-mass rescaling.

\begin{algorithm}[h]
\small
\caption{Dynamic-Prefix OPD Truncation}
\label{alg:dynamic_prefix_opd}
\begin{algorithmic}[1]
\Require Response tokens \(y_{1:T}\), token advantages \(A_{1:T}\), loss mask \(l_{1:T}\), teacher probabilities on student top-\(K\) candidates, NLTK sentence segments \(\mathcal{B}_1,\ldots,\mathcal{B}_n\)
\Ensure Truncated and rescaled advantages \(A'_{1:T}\)
\For{each response token position \(t\)}
    \State \(\mathcal{C}_t \gets \operatorname{TopK}_{a}\,p_S(a\mid x,y_{<t})\)
    \State Rank candidates in \(\mathcal{C}_t\) by teacher log-probability
    \State \(a_t^{(1)} \gets \arg\max_{a\in\mathcal{C}_t}\log p_T(a\mid x,y_{<t})\)
    \State \(a_t^{(2)} \gets \arg\max_{a\in\mathcal{C}_t\setminus\{a_t^{(1)}\}}\log p_T(a\mid x,y_{<t})\)
    \State \(M_t \gets \log p_T(a_t^{(1)}\mid x,y_{<t})-\log p_T(a_t^{(2)}\mid x,y_{<t})\)
\EndFor
\For{each sentence segment \(\mathcal{B}_i\)}
    \State \(S_i \gets \log\!\left(1+\frac{1}{|\mathcal{B}_i|}\sum_{t\in\mathcal{B}_i}M_t\right)\)
\EndFor
\State \(\operatorname{BIC}_0 \gets n\log((\sum_i(S_i-\bar S)^2+\epsilon)/n)+\log n\)
\State \(\mathcal{T}\gets\{\tau\in\{1,\ldots,n-1\}:\bar S_{>\tau}<\bar S_{\le\tau}\}\)
\For{each \(\tau\in\mathcal{T}\)}
    \State \(\operatorname{RSS}_1(\tau)\gets \sum_{i\le\tau}(S_i-\bar S_{\le\tau})^2+\sum_{i>\tau}(S_i-\bar S_{>\tau})^2\)
    \State \(\operatorname{BIC}_1(\tau)\gets n\log((\operatorname{RSS}_1(\tau)+\epsilon)/n)+3\log n\)
\EndFor
\State \(\tau^\star \gets \operatorname*{arg\,min}_{\tau\in\mathcal{T}}\operatorname{BIC}_1(\tau)\)
\If{\(\mathcal{T}\neq\emptyset\) and \(\operatorname{BIC}_1(\tau^\star)<\operatorname{BIC}_0\)}
    \State \(q_t\gets 1\) for tokens before the boundary after segment \(\tau^\star\), and \(q_t\gets 0\) otherwise
\Else
    \State \(q_t\gets 1\) for all valid response tokens
\EndIf
\State \(A'_t \gets A_t\,q_t\,\frac{\sum_t l_t}{\max(\sum_t l_tq_t,\epsilon)}\)
\State \Return \(A'_{1:T}\)
\end{algorithmic}
\end{algorithm}

\subsection{Theoretical Motivation for Local Teachability}
\label{sec:appendix_theoretical_motivation}

This section gives a lightweight theoretical motivation for the release rule. The goal is not to prove that the BIC criterion always identifies a true change point, nor that truncation is universally variance-reducing. Instead, we formalize why local discriminability of teacher feedback is relevant to OPD, and why a penalized downward-change rule is a statistically meaningful way to operationalize this signal.

\paragraph{Two notions of advantage variation.}
The analysis in Section~\ref{sec:analysis} uses the dispersion of sampled-token advantages,
\[
A_t(y_t)=\log p_T(y_t\mid x,y_{<t})-\log p_S(y_t\mid x,y_{<t}),
\]
along a realized trajectory. This statistic is a diagnostic symptom: it shows that dense OPD feedback can lose token-specific variation as the response progresses. The release rule does not directly use this trajectory-wise \(\operatorname{std}(A_t(y_t))\). Instead, it uses a state-local proxy computed at a fixed prefix \(h_t=(x,y_{<t})\): whether the teacher can still distinguish among the student's likely next-token candidates. The derivation below is about this second object, namely action-wise variation inside the student's local candidate set \(\mathcal C_t\). Thus, sampled-path advantage dispersion motivates the problem, while the top-\(K\) teacher margin operationalizes the release decision.

\paragraph{Local teacher signal and policy-gradient usefulness.}
For a candidate next token \(a\) at a fixed prefix \(h_t\), define the action-wise OPD advantage
\[
A_t(a)=\log p_T(a\mid h_t)-\log p_S(a\mid h_t),
\]
and the score vector
\[
s_t(a)=\nabla_\theta \log p_S(a\mid h_t).
\]
The local expected OPD update over a candidate distribution \(\pi_t\) supported on the student's reachable set \(\mathcal C_t\) is
\[
g_t
=
\mathbb{E}_{a\sim \pi_t}
\left[
A_t(a)s_t(a)
\right].
\]
This is an action-wise analysis at one prefix, not a temporal statistic over adjacent sampled tokens. In the standard policy-gradient baseline argument, if the expectation is taken under the full student distribution, then \(\mathbb{E}_{a\sim p_S(\cdot\mid h_t)}[s_t(a)]=0\), so subtracting any action-independent baseline from \(A_t(a)\) does not change the expected update. For the restricted top-\(K\) set, this identity is only approximate because \(\pi_t\) is a locally renormalized proxy; equivalently, we can view the following centered expression as the informative component that removes action-independent offsets:
\[
\tilde g_t
\approx
\mathbb{E}_{a\sim \pi_t}
\left[
(A_t(a)-\bar A_t)s_t(a)
\right],
\qquad
\bar A_t=\mathbb{E}_{a\sim \pi_t}[A_t(a)].
\]
By Cauchy's inequality,
\[
\|\tilde g_t\|^2
\le
\operatorname{Var}_{a\sim \pi_t}(A_t(a))
\cdot
\mathbb{E}_{a\sim \pi_t}\|s_t(a)\|^2 .
\]
This bound explains the central distinction made in the paper. A nonzero sampled-token advantage along a trajectory does not by itself imply a useful dense supervision signal. What matters for local teaching is whether the teacher creates action-dependent contrast among alternatives the student is likely to take. If the teacher's preference over the student's reachable alternatives becomes locally flat, then the action-dependent component of the update becomes small, and dense token-level feedback becomes less informative.

\paragraph{Why teacher margin is a proxy for teachability.}
One could also compute \(\operatorname{Var}_{\mathcal C_t}(A_t(a))\) directly over the same top-\(K\) candidate set. This would measure action-wise advantage dispersion within the local support, and its computational cost is comparable when teacher and student scores for all top-\(K\) candidates are already available. We use the teacher's nearest-competitor margin instead because it targets a narrower decision question: whether the teacher has a clear first choice over the actions the student is most likely to take. The margin is defined inside the student's top-\(K\) candidate set as
\[
M_t
=
\log p_T(a_t^{(1)}\mid h_t)
-
\log p_T(a_t^{(2)}\mid h_t),
\]
where \(a_t^{(1)}\) and \(a_t^{(2)}\) are the teacher's top two candidates within \(\mathcal C_t\). This quantity does not fully characterize the local advantage distribution. Instead, it measures a narrower but useful property: whether the teacher can still express a sharp preference among actions the student is likely to take. A sustained decrease in \(M_t\) therefore indicates that teacher feedback may remain globally different from the student while becoming less locally actionable.

\paragraph{SNR view of release.}
Let \(P\) denote the prefix retained by the release rule and \(R\) denote the released suffix. Here the segment-level signal is understood through the informative action-dependent component \(\tilde g_t\), whose magnitude is controlled by local advantage variation in the preceding bound. For a fixed update direction \(u\), define the directional signal and variance
\[
m_P = u^\top \mathbb{E}\!\left[\sum_{t\in P} \tilde g_t\right],
\qquad
v_P = \operatorname{Var}\!\left(u^\top \sum_{t\in P} \tilde g_t\right),
\]
and analogously \(m_R,v_R\) for the suffix. The full OPD update has directional SNR
\[
\mathrm{SNR}_{\mathrm{full}}
=
\frac{(m_P+m_R)^2}{v_P+v_R},
\]
whereas the released update, up to the loss-mass rescaling used in the paper, has the same directional SNR as the retained prefix:
\[
\mathrm{SNR}_{\mathrm{release}}
=
\frac{m_P^2}{v_P}.
\]
Therefore release improves directional SNR whenever
\[
\frac{m_P^2}{v_P}
\ge
\frac{(m_P+m_R)^2}{v_P+v_R},
\]
or equivalently,
\[
\frac{v_R}{v_P}
\ge
2\frac{m_R}{m_P}
+
\left(\frac{m_R}{m_P}\right)^2 .
\]
This condition captures the intended regime of local teachability collapse: the suffix contributes little aligned teacher signal relative to its variance. In that case, continuing dense OPD supervision on the suffix can lower the effective SNR of the update. The release rule is designed to detect this regime using the teacher-margin curve.

\paragraph{Role of the BIC-style criterion.}
The segment scores \(S_i\) are noisy summaries of local teacher contrast. The BIC-style rule compares a no-change model,
\[
S_i=\mu+\epsilon_i,
\]
against a one-drop model,
\[
S_i=
\begin{cases}
\mu_{\mathrm{pre}}+\epsilon_i, & i\le \tau,\\
\mu_{\mathrm{post}}+\epsilon_i, & i>\tau,
\end{cases}
\qquad
\mu_{\mathrm{pre}}>\mu_{\mathrm{post}}.
\]
The profiled RSS-BIC objective penalizes the additional parameters of the one-drop model:
\[
\operatorname{BIC}_0
=
n\log\left(\frac{\operatorname{RSS}_0+\epsilon}{n}\right)
+
\log n,
\]
\[
\operatorname{BIC}_1(\tau)
=
n\log\left(\frac{\operatorname{RSS}_1(\tau)+\epsilon}{n}\right)
+
3\log n.
\]
Thus, a release is accepted only when the reduction in residual error is large enough to overcome the complexity penalty. This does not guarantee recovery of a true latent change point for every finite trajectory. Rather, it makes the rule statistically disciplined: it avoids releasing merely because of small random fluctuations, while still allowing each trajectory to choose its own supervision boundary.

\begin{wrapfigure}[18]{r}{0.60\columnwidth}
    \centering
    \includegraphics[width=0.4\columnwidth]{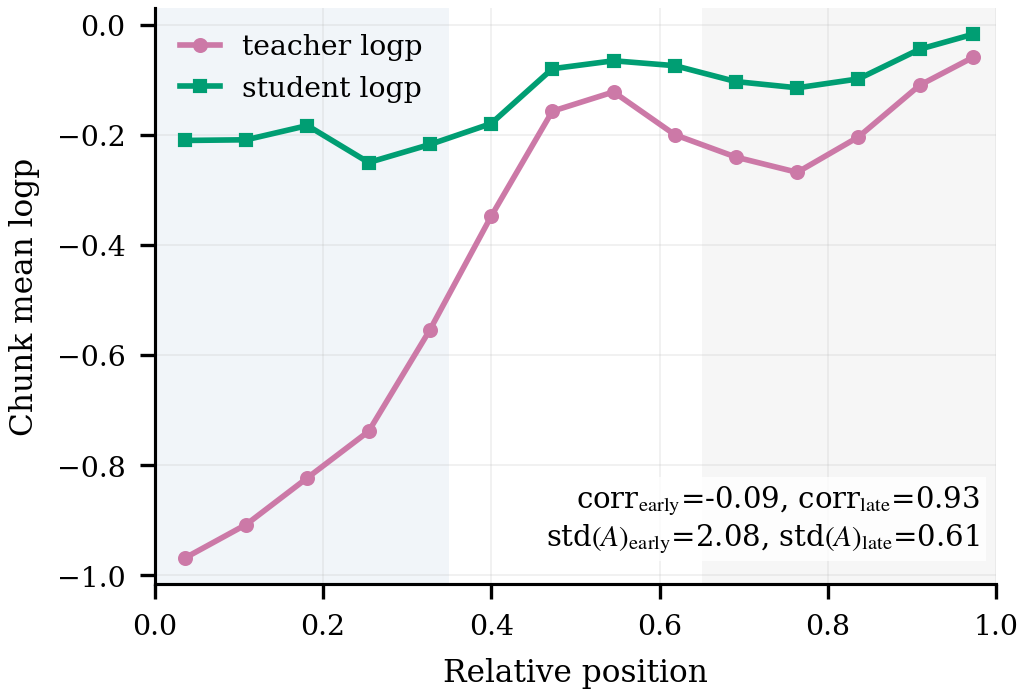}
    \caption{\textbf{Qualitative Example}: A representative rollout shows that while teacher and student log-probabilities diverge early on, they become highly aligned in later stages (log-probabilities correlation increases from -0.09 to 0.93). The resulting reduction in advantage variation ($\text{std}(A)$ drops from 2.08 to 0.61) renders the dense feedback less actionable for guiding specific local policy improvements.}
    \label{fig:appendix_local_teachability_case}
\end{wrapfigure}

\paragraph{Interpretation.}
The analysis supports the main claim of the paper: dense OPD should depend not only on whether a teacher-student gap exists, but also on whether the teacher's feedback remains locally discriminative over actions the student can plausibly take. The proposed release rule operationalizes this principle by using a state-local teacher margin as a teachability proxy and a penalized downward-change criterion to decide when dense supervision should stop.

\subsection{Additional Diagnostic}
\label{sec:appendix_additional_diagnostics}

Figure~\ref{fig:appendix_local_teachability_case} illustrates this symptom. Early in the rollout, the teacher and student log-probability curves diverge substantially, yielding highly variable advantage values. Yet, as the generation prefix $(x, y_{<t})$ lengthens, the predictions of both models become heavily conditioned on the same accumulated context. Their sampled-token log-probability trajectories align, the standard deviation of the advantages drops, and the scalar feedback becomes less informative as a token-specific priority signal.

\begin{figure}[h]
    \centering
    \begin{minipage}{0.99\columnwidth}
        \centering
        \includegraphics[width=0.98\linewidth]{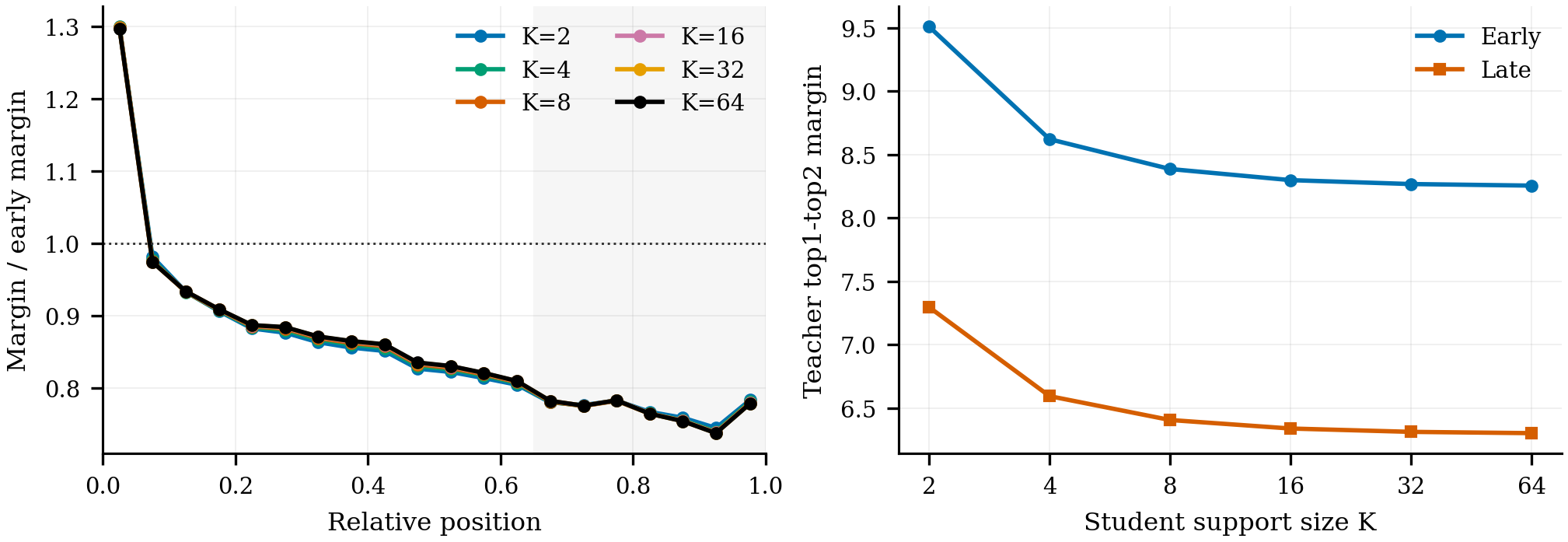}
        \caption{\textbf{Support-size sensitivity.} Left: The normalized teacher-margin curves remain similar across \(K\in\{2,4,8,16,32,64\}\), supporting the robustness of the local-contrast decline. Right: Early segments consistently exhibit larger teacher top-1--top-2 margins than late segments across support sizes, indicating that the decline in local teachability is not an artifact of a particular \(K\).}
        \label{fig:topk_support_margin_sensitivity}
    \end{minipage}
\end{figure}
Figure~\ref{fig:topk_support_margin_sensitivity} checks whether the normalized margin decline depends strongly on the chosen support size. The curves remain similar across \(K\in\{2,4,8,16,32,64\}\), which supports the interpretation that the observed decay is a property of teacher ranking sharpness over student-reachable actions rather than an artifact of one support size. The main experiments fix the margin definition to top1--top2 and report the training-level support-size check in Table~\ref{tab:support_size_adv_check}.

\begin{wrapfigure}[18]{r}{0.50\columnwidth}
    \centering
    \includegraphics[width=0.9\linewidth]{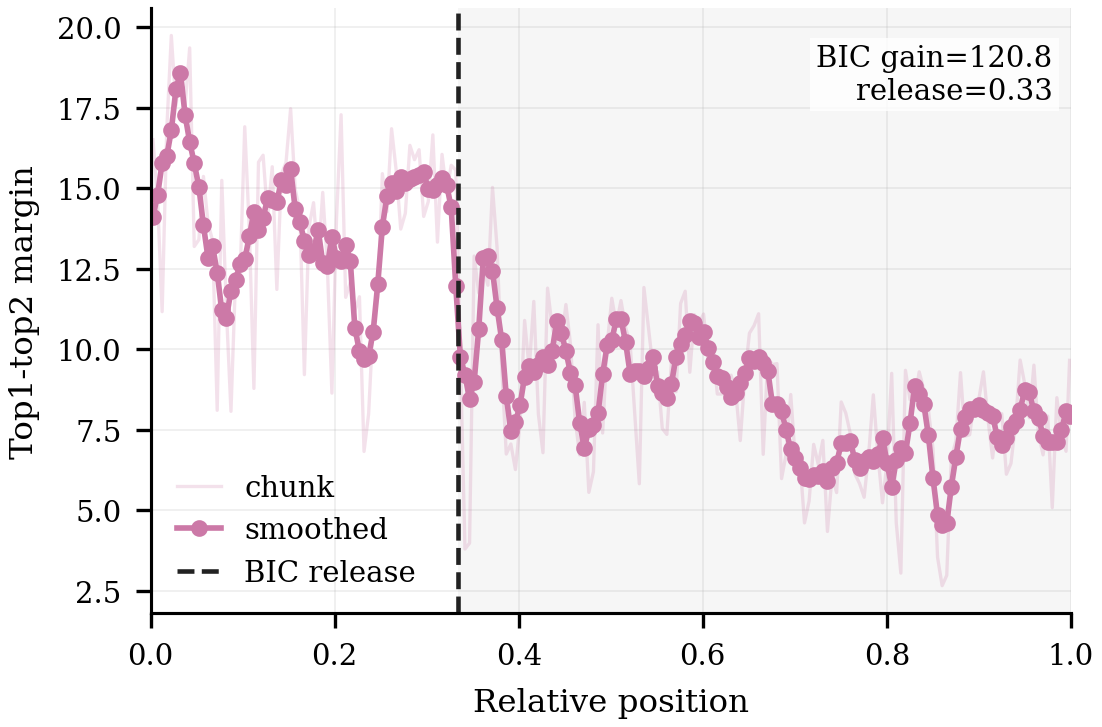}
    \caption{\textbf{Example BIC-style release.} A representative rollout shows an accepted downward change point in the teacher top1--top2 margin over the student's top-\(K\) candidates.}
    \label{fig:method_release_case}
\end{wrapfigure}

Figure~\ref{fig:method_release_case} shows how the BIC-style rule behaves on one trajectory. The accepted release point is not meant to prove that the subsequent tokens are useless; rather, it marks where the aggregated margin curve is better summarized by a lower-contrast post-change segment. This example complements the distribution-level evidence in the main text by showing the object on which the rule operates. The same implementation is applied to all rollouts, with no benchmark-specific cutoff or per-sample manual adjustment. If the fixed criterion does not accept a downward change point, the trajectory falls back to full OPD supervision.

\begin{wrapfigure}[20]{r}{0.50\columnwidth}
    \centering
        \includegraphics[width=0.98\linewidth]{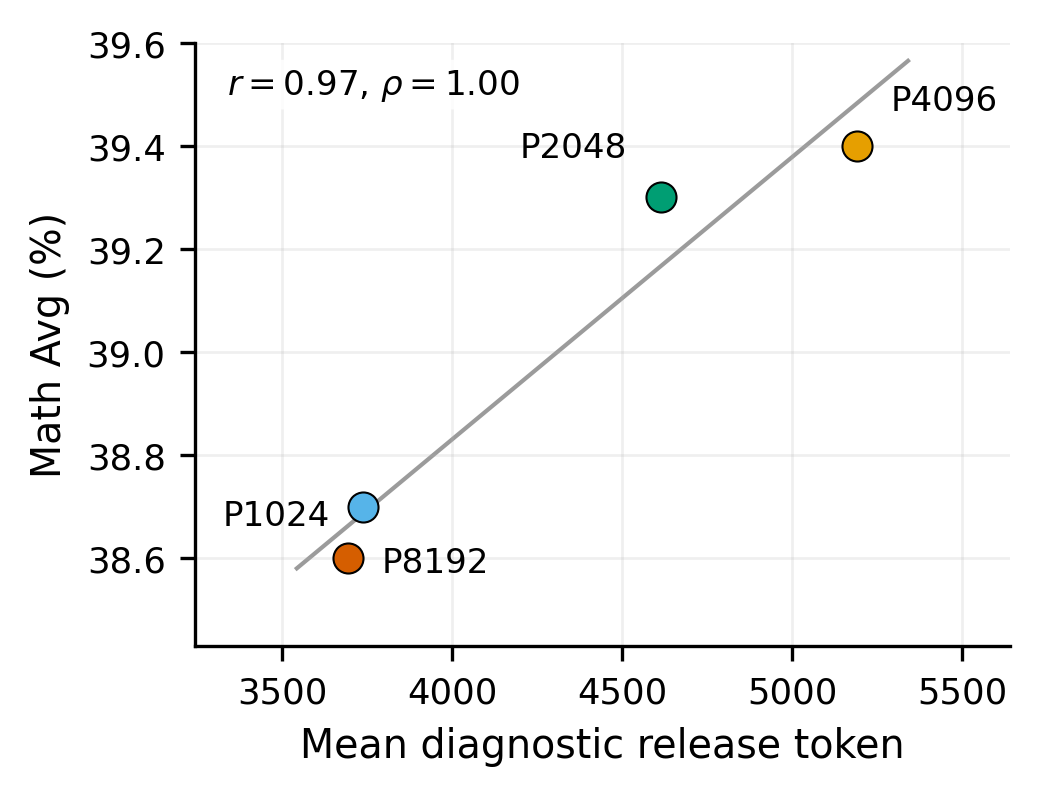}
        \caption{\textbf{Diagnostic release positions and fixed-prefix performance.} Each point is one fixed-prefix checkpoint; with only four checkpoints, this is descriptive rather than a correlation claim.}
        \label{fig:prefix_release_token_vs_avg}
\end{wrapfigure}

Figure \ref{fig:prefix_release_token_vs_avg} illustrates the relationship between fixed-prefix checkpoints and the diagnostic release positions estimated from their respective rollouts. Each data point represents a checkpoint trained with a specific global prefix cutoff. The horizontal axis denotes the mean diagnostic release token, calculated by applying our teacher-margin release rule to rollouts from that checkpoint. The vertical axis represents the checkpoint's performance averaged across five downstream math benchmarks. Consequently, this visualization evaluates whether a checkpoint’s rollout distribution maintains locally teachable regions throughout the response, rather than assessing the choice of prefix length in isolation.

This trend suggests a mechanistic interpretation. Better fixed-prefix checkpoints tend to have rollouts whose margin curves imply later diagnostic release positions, meaning that the teacher remains locally discriminative for a longer portion of those trajectories. This suggests that a good fixed prefix is not good merely because a particular token index is intrinsically special. Rather, it works when that global cutoff happens to retain a region where the teacher still provides useful local contrast. Conversely, if the local-teachability curve collapses early, extending dense supervision by a larger fixed prefix may add low-value suffix tokens. Because there are only four fixed-prefix checkpoints, this plot should not be read as a statistical correlation claim. Its role is to illustrate why a single nominal prefix length is an incomplete explanation: the useful supervision boundary depends on the trajectory-level teacher signal as well as the global cutoff.

\subsection{Release-Point Case Studies}
\label{sec:appendix_release_case_study}

The following case studies illustrate the specific behaviors that the release rule is designed to exclude from supervision. These instances do not constitute inference-time truncation. Because the trajectory-specific rule exclusively modifies the training objective, the model remains capable of generating complete responses during evaluation.

While the blue regions are not guaranteed to be error-free, they encompass the primary reasoning or setup preceding the release point. In contrast, the red regions highlight repetitive or noisy continuations that would otherwise receive potentially counterproductive dense supervision from the teacher model. These examples thus demonstrate why dense OPD supervision on terminal regions can be of limited utility.

\paragraph{Case 1: repeated low-value continuation after the release point.}
The first example asks for \(\lim_{n\to\infty} n^3x^n(1-x)^4\) for \(x\in[0,1]\). The pre-release prefix already contains the correct case split and the central observation that exponential decay dominates polynomial growth for \(0<x<1\). After the BIC release point, the continuation repeatedly restates the same argument and includes several false starts.

\begin{tcolorbox}[
  enhanced,
  breakable,
  colback=blue!4,
  colframe=blue!60!black,
  boxrule=0.7pt,
  arc=1.2mm,
  left=1.4mm,
  right=1.4mm,
  top=1.0mm,
  bottom=1.0mm,
  title=\textbf{Before release: core setup before noisy continuation},
  fonttitle=\small,
  fontupper=\scriptsize,
]

We are asked to evaluate the limit:

$$
\lim_{n \to \infty} n^3 x^n (1 - x)^4 \quad \text{for } x \in [0, 1].
$$

We will analyze this limit by considering different cases based on the value of $ x $ in the interval $ [0, 1] $.

---

\#\#\# Step 1: Understanding the Behavior

Let's denote:
$$
a_n = n^3 x^n (1 - x)^4
$$
We are to evaluate:
$$
\lim_{n \to \infty} a_n
$$

We'll analyze this limit depending on whether $ x < 1 $, $ x = 1 $, or $ x = 0 $.

---

\#\#\# Case 1: $ x = 0 $

Then $ x^n = 0^n = 0 $ for $ n \geq 1 $, so:
$$
a_n = n^3 \cdot 0 = 0 \quad \text{for all } n \geq 1
$$
Therefore:
$$
\lim_{n \to \infty} a_n = 0
$$

---

\#\#\# Case 2: $ x = 1 $

Then $ x^n = 1^n = 1 $, so:
$$
a_n = n^3 \cdot 1 \cdot (1 - 1)^4 = n^3 \cdot 1 \cdot 0 = 0
$$
So again:
$$
\lim_{n \to \infty} a_n = 0
$$

---

\#\#\# Case 3: $ 0 < x < 1 $

Now $ x \in (0, 1) $, so $ x^n \to 0 $ as $ n \to \infty $, but $ n^3 $ grows to infinity. We need to understand the \textbf{rate} at which $ x^n $ decays.

Recall that for fixed $ x \in (0,1) $, $ x^n \to 0 $ exponentially fast.

We can compare $ n^3 x^n $ with its exponential decay.

Let's consider the ratio:
$$
\frac{a_{n+1}}{a_n} = \frac{(n+1)^3 x^{n+1} (1 - x)^4}{n^3 x^n (1 - x)^4} = \left(\frac{n+1}{n}\right)^3 x
$$
$$
= \left(1 + \frac{1}{n}\right)^3 x
$$

As $ n \to \infty $, $ \left(1 + \frac{1}{n}\right)^3 \to 1 $, so the ratio $ \frac{a_{n+1}}{a_n} \to x $. Since $ x \in (0,1) $, this ratio approaches a value less than 1.

This suggests that $ a_n \to 0 $, as the sequence $ a_n $ is decreasing and bounded below by 0 (since $ x > 0 $, $ x^n > 0 $), and the ratio is less than 1.

But let's be more precise.

\end{tcolorbox}

\begin{tcolorbox}[
  enhanced,
  breakable,
  colback=red!4,
  colframe=red!55!black,
  boxrule=0.7pt,
  arc=1.2mm,
  left=1.4mm,
  right=1.4mm,
  top=1.0mm,
  bottom=1.0mm,
  title=\textbf{After release: low-value content},
  fonttitle=\small,
  fontupper=\scriptsize,
]

We can use the \textbf{standard result}:

For $ 0 < x < 1 $, the sequence $ n^k x^n \to 0 $ as $ n \to \infty $ for any fixed $ k \in \mathbb{R} $.

This is a well-known fact in analysis: exponential decay dominates polynomial growth.

In fact, we can apply the \textbf{exponential decay} bound:

\[
x^n \leq e^{-n \cdot \frac{\ln x}{1}} \quad \text{since } x = e^{\ln x} \text{ and } \ln x < 0 \Rightarrow -\ln x > 0
\]

But let's use a more precise method.

Let's use the \textbf{inequality}:
\[
\forall n \geq 1, \quad x^n \leq \frac{1}{n^k} \quad \text{?}
\]
No, that's not helpful.

Instead, we use:
\[
\lim_{n \to \infty} n^3 x^n (1 - x)^4 = 0 \quad \text{for } 0 < x < 1
\]

Why?

Because $ x^n \to 0 $ exponentially, while $ n^3 $ grows polynomially. The \textbf{exponential decay dominates polynomial growth}, so the product goes to 0.

Let's make a more formal argument.

Let $ x \in (0,1) $. Then $ x^n \to 0 $, and $ n^3 $ grows, but we can write:
\[
n^3 x^n = n^3 e^{n \ln x} = e^{n \ln n^3 + n \ln x} = e^{n (\ln n^3 + \ln x)}
\]
So:
\[
\lim_{n \to \infty} n^3 x^n = \lim_{n \to \infty} e^{n (\ln n^3 + \ln x)} = \infty \cdot e^{-\infty} \quad \text{?}
\]
Wait --- no, wait.

Actually, $ n \ln n^3 = 3 n \ln n $, and $ n \ln x $ is negative since $ x < 1 $, so $ n \ln x \to -\infty $.

So:
\[
n^3 x^n = e^{n \ln n^3 + n \ln x} = e^{n (3 \ln n + \ln x)}
\]
We analyze the exponent:
\[
n (3 \ln n + \ln x) \to \infty \text{ because } n \ln n \to \infty \text{ and } \ln x < 0, \text{ so } n \ln x \to -\infty \text{ but much slower than } n \ln n
\]
So overall, $ n^3 x^n \to \infty \cdot e^{-\infty} $ --- is that 0?

Actually, $ n \ln n \to \infty $, and $ n \ln x \to -\infty $, but the exponent grows to $ \infty $ because $ n \ln n \to \infty $, so $ n^3 x^n \to 0 $ \textbf{exponentially faster than any polynomial growth}.

Wait --- let's make a \textbf{comparison}:

We can use:
\[
n^3 x^n = e^{n \ln n^3 + n \ln x} = e^{n(3 \ln n + \ln x)}
\]
Let $ a_n = n^3 x^n $, so $ \ln a_n = n (3 \ln n + \ln x) $

Now analyze $ \ln a_n $:
\begin{itemize}
    \item $ 3 \ln n \to \infty $
    \item $ \ln x < 0 $, so $ n \ln x \to -\infty $
    \item So $ \ln a_n \to \infty $ \textbf{if} $ n \ln n \to \infty $ --- which it does.
\end{itemize}

But is it $ \infty $? Yes, because $ 3 \ln n \to \infty $, so the entire exponent $ n (3 \ln n + \ln x) \to \infty $, because $ n \ln n \to \infty $, and $ \ln x $ is fixed.

But $ \ln x $ is negative, so the exponent is $ \infty $, \textbf{but the exponent grows to $ \infty $}, so $ e^{n(3 \ln n + \ln x)} \to \infty $? Wait, no!

Actually, it's:
\[
\lim_{n \to \infty} e^{n(3 \ln n + \ln x)} = \lim_{n \to \infty} e^{3 n \ln n + n \ln x} = \lim_{n \to \infty} n^{3n} e^{n \ln x}
\]
But $ e^{n \ln x} = x^n $, so we are back.

But now, we can write:
\[
n^3 x^n = e^{n \ln n^3 + n \ln x} = e^{n (3 \ln n + \ln x)} = e^{n \ln n^3} \cdot e^{n \ln x} = (n^3)^n \cdot x^n
\]
Wait --- that's not helpful.

But the key idea is that:
\begin{itemize}
    \item $ n^3 \to \infty $
    \item $ x^n \to 0 $ exponentially
    \item But since $ x^n $ decays faster than any polynomial, the product $ n^3 x^n \to 0 $.
\end{itemize}

This is a standard result.

\textbf{Standard limit:}
\[
\lim_{n \to \infty} n^k x^n = 0 \quad \text{for } x \in (0,1), \text{ any } k > 0
\]

To see this, consider that:
\[
x^n = e^{n \ln x} \to 0 \text{ exponentially fast}
\]
and $ n^k $ grows polynomially, so exponential decay dominates polynomial growth.

Thus, for fixed $ k $, we have:
\[
\lim_{n \to \infty} n^k x^n = 0
\]

In our case, $ k = 3 $, so for $ x \in (0,1) $, the limit is 0.

But wait: is that true?

Yes --- for example, take $ x = 1/2 $, then $ (1/2)^n $ decays exponentially, so $ n^3 (1/2)^n \to 0 $.

Similarly, $ x = 1/10 $: $ (1/10)^n $ decays faster than $ n^3 $.

So yes, the limit is 0.

---

\#\#\# \textbf{Case 4: $ x = 0 $ or $ x = 1 $: already handled}

Both give 0.

---

\#\#\# \textbf{Special Case: $ x = 0.5 $}

Compute $ n^3 (0.5)^n $: it's clearly going to 0.

---

\#\#\# \textbf{Conclusion}

\begin{itemize}
    \item If $ x \in (0,1) $: $ x^n \to 0 $ exponentially, $ n^3 $ polynomially, so $ n^3 x^n \to 0 $
    \item If $ x = 0 $: 0
    \item If $ x = 1 $: 0
\end{itemize}

Thus, in all cases $ x \in [0,1] $, the limit is:

\[
\boxed{0}
\]

--- 

\#\#\# $\checkmark$ \textbf{Final Answer:}
\[
\boxed{0}
\]

\end{tcolorbox}

\paragraph{Case 2: correct conclusion with noisy self-correction.}
The second example asks for \(\lim_{n\to\infty}P(\sum_{i=1}^n-\ln(1-X_i)\ge n)\). The pre-release region contains the main reduction toward a sum of transformed i.i.d. variables and the intended central-limit-theorem route. The released continuation repeatedly debates whether a Gamma distribution is symmetric, including erroneous intermediate statements, before returning to the same limiting answer.

\begin{tcolorbox}[
  enhanced,
  breakable,
  colback=blue!4,
  colframe=blue!60!black,
  boxrule=0.7pt,
  arc=1.2mm,
  left=1.4mm,
  right=1.4mm,
  top=1.0mm,
  bottom=1.0mm,
  title=\textbf{Before release: core setup before noisy continuation},
  fonttitle=\small,
  fontupper=\scriptsize,
]

We are given a sequence of i.i.d. uniform $(0,1)$ random variables $X_1, X_2, X_3, \ldots$, and we are to compute:

\[
\lim_{n \to \infty} P\left(-\ln(1 - X_1) - \ln(1 - X_2) - \cdots - \ln(1 - X_n) \geq n\right)
\]

\vspace{1em}
\noindent ---
\vspace{1em}

\noindent \#\#\# Step 1: Understand the expression inside the probability

Let:
\[
S_n = -\sum_{i=1}^n \ln(1 - X_i)
\]
We are to compute:
\[
\lim_{n \to \infty} P(S_n \geq n)
\]

We observe that $X_i \sim \text{Uniform}(0,1)$, so $1 - X_i \sim \text{Uniform}(0,1)$, and $\ln(1 - X_i)$ is a decreasing function. Since $X_i \in (0,1)$, $\ln(1 - X_i) \in (-\infty, 0)$, so $-\ln(1 - X_i) \in (0, \infty)$.

Now consider:
\[
-\ln(1 - X_i) = \ln\left(\frac{1}{1 - X_i}\right)
\]

So:
\[
S_n = \sum_{i=1}^n \ln\left(\frac{1}{1 - X_i}\right)
\]

Let’s define:
\[
Y_i = \ln\left(\frac{1}{1 - X_i}\right)
\]
So $Y_i = -\ln(1 - X_i)$, and we want:
\[
P(S_n \geq n) = P\left( \sum_{i=1}^n Y_i \geq n \right)
\]

\vspace{1em}
\noindent ---
\vspace{1em}

\noindent \#\#\# Step 2: Analyze the distribution of $Y_i$

Since $X_i \sim \text{Uniform}(0,1)$, we have:
\[
1 - X_i \sim \text{Uniform}(0,1), \quad \text{so } \frac{1}{1 - X_i} \sim \text{Uniform}(1, \infty)
\]
But more importantly, $Y_i = \ln\left(\frac{1}{1 - X_i}\right) = -\ln(1 - X_i)$

Let’s compute the cumulative distribution function (CDF) of $Y_i$.

Let $F_{Y_i}(y) = P(Y_i \leq y) = P\left(-\ln(1 - X_i) \leq y\right)$

Solve:
\[
-\ln(1 - X_i) \leq y \quad \Rightarrow \quad \ln(1 - X_i) \geq -y \quad \Rightarrow \quad 1 - X_i \geq e^{-y} \quad \Rightarrow \quad X_i \leq 1 - e^{-y}
\]

So:
\[
F_{Y_i}(y) = P(X_i \leq 1 - e^{-y}) = 1 - e^{-y}, \quad \text{for } y \geq 0
\]

Therefore:
\[
Y_i \sim \text{Exponential}(1), \quad \text{since } F_{Y_i}(y) = 1 - e^{-y}, \quad \text{and CDF of Exp}(1) \text{ is } 1 - e^{-y}
\]

So:
\[
Y_i \sim \text{Exp}(1)
\]

That is, $Y_i$ has an exponential distribution with rate parameter 1.

\vspace{1em}
\noindent ---
\vspace{1em}

\noindent \#\#\# Step 3: Use independence

Since $X_i$ are i.i.d., $Y_i = -\ln(1 - X_i)$ are i.i.d. exponential random variables with $\lambda = 1$.

Thus, $Y_1, Y_2, \ldots, Y_n$ are i.i.d. $ \text{Exp}(1) $.

We are to compute:
\[
\lim_{n \to \infty} P\left( \sum_{i=1}^n Y_i \geq n \right)
\]

This is a sum of i.i.d. exponential random variables with $\lambda = 1$, and we are asking about the probability that the sum is at least $n$.

Let’s denote:
\[
T_n = \sum_{i=1}^n Y_i
\]

We are to compute:
\[
\lim_{n \to \infty} P(T_n \geq n)
\]

\vspace{1em}
\noindent ---
\vspace{1em}

\noindent \#\#\# Step 4: Use the Central Limit Theorem (CLT)

Since $Y_i \sim \text{Exp}(1)$, they are \textbf{light-tailed} and we can apply the \textbf{Central Limit Theorem}.

Recall that for $Y_i \sim \text{Exp}(1)$, the mean and variance are:
\[
\mathbb{E}[Y_i] = 1, \quad \text{Var}(Y_i) = 1
\]

So:
\[
\mathbb{E}[T_n] = n \cdot 1 = n
\]
\[
\text{Var}(T_n) = n \cdot 1 = n
\]
\[
\text{SD}(T_n) = \sqrt{n}
\]

Therefore, $T_n$ has mean $n$ and standard deviation $\sqrt{n}$.

Now, the random variable $T_n - n$ has mean 0 and variance $n$.

We now consider:
\[
P(T_n \geq n) = P\left( T_n - n \geq 0 \right)
\]

Let $Z_n = \frac{T_n - n}{\sqrt{n}}$, then $Z_n \to \mathcal{N}(0,1)$ in distribution as $n \to \infty$, by the CLT.

So:
\[
P(T_n \geq n) = P\left( Z_n \geq 0 \right) \to \frac{1}{2}
\]

Because the standard normal distribution is symmetric around 0, and $P(Z \geq 0) = 1/2$.

But wait — is this accurate?

Let me double-check: For $Z_n = \frac{T_n - n}{\sqrt{n}} \to \mathcal{N}(0,1)$, then:

\[
P(T_n \geq n) = P\left( Z_n \geq 0 \right) \to \frac{1}{2}
\]

Yes.

But we must be careful:

\end{tcolorbox}

\begin{tcolorbox}[
  enhanced,
  breakable,
  colback=red!4,
  colframe=red!55!black,
  boxrule=0.7pt,
  arc=1.2mm,
  left=1.4mm,
  right=1.4mm,
  top=1.0mm,
  bottom=1.0mm,
  title=\textbf{After release: noisy continuation content},
  fonttitle=\small,
  fontupper=\scriptsize,
]

the CLT gives that $T_n$ is asymptotically normal with mean $n$ and variance $n$, so $T_n - n$ is normal with mean 0 and variance $n$, so $Z_n = \frac{T_n - n}{\sqrt{n}}$ is standard normal.

Therefore, $P(T_n \geq n) = P(Z_n \geq 0) \to \frac{1}{2}$

But wait --- is this correct?

Let's think about the \textbf{simulation}. For large $n$, $T_n \sim \text{Normal}(n, n)$ --- so it's a normal variable with mean $n$ and variance $n$. The distribution is very skewed? Wait --- no, \textbf{exponential} variables have a \textbf{heavy tail}, but the sum of them is \textbf{approximately normal}.

But yes, the \textbf{Central Limit Theorem} says that the sum of i.i.d. random variables with finite variance, \textbf{even if they aren't symmetric}, will converge in distribution to a normal distribution with mean equal to the sum of means and variance equal to the sum of variances.

So for $Y_i \sim \text{Exp}(1)$, the CLT applies:

\begin{itemize}
    \item $\mathbb{E}[Y_i] = 1$
    \item $\text{Var}(Y_i) = 1$
\end{itemize}

So for $T_n = \sum Y_i$, we have:
$$
\mathbb{E}[T_n] = n, \quad \text{Var}(T_n) = n, \quad \text{SD} = \sqrt{n}
$$

And $T_n$ is the sum of i.i.d. exponential(1) variables.

So $T_n \sim N(n, n)$ in distribution for large $n$ --- but \textbf{we are not taking the limit} of the distribution, but computing the \textbf{probability} that $T_n \geq n$.

We ask: as $n \to \infty$, what happens to $P(T_n \geq n)$?

Let's consider a \textbf{realistic} case: Let's consider a standard normal $Z \sim N(0,1)$. Then $P(Z \geq 0) = 0.5$.

Now, $T_n \sim N(n, n)$ --- this means that $T_n$ is centered at $n$ with standard deviation $\sqrt{n}$. So $T_n$ is concentrated around $n$, and the deviation from $n$ increases with $n$.

But the question is about $P(T_n \geq n)$.

Let's define $Z_n = \frac{T_n - n}{\sqrt{n}}$, then:

$$
P(T_n \geq n) = P\left( Z_n \geq 0 \right) \to \frac{1}{2}
$$

Because $Z_n \to \mathcal{N}(0,1)$ in distribution, and $P(Z \geq 0) = 1/2$.

Therefore, the limit is:
$$
\lim_{n \to \infty} P(T_n \geq n) = \frac{1}{2}
$$

But wait --- is there a \textbf{finite} limit?

Let me test with \textbf{large $n$}: Suppose $n$ is large, then $T_n \sim N(n, n)$. So the probability that $T_n \geq n$ is approximately $P(Z \geq 0) = 1/2$, since the normal distribution is symmetric.

However, the \textbf{KL divergence} or \textbf{convergence} is just a point-wise limit: since $T_n$ is \textbf{approximately normal}, then $P(T_n \geq n)$ tends to $1/2$.

But wait --- is this correct?

Wait --- let's think about the \textbf{variance}. $T_n$ has mean $n$, and variance $n$, so $T_n - n$ has variance $n$, which is growing as $n \to \infty$. So $T_n$ is very far from $n$ in \textbf{terms of standard deviation}?

No, \textbf{standard deviation} is $\sqrt{n}$, so $T_n$ is centered at $n$, and has standard deviation $\sqrt{n}$, so it's \textbf{not} growing with $n$, but the \textbf{deviation} from the mean is growing as $\sqrt{n}$. But we are checking the probability that it's greater or equal to $n$.

Let's make a \textbf{change of variables}:

Let $T_n = n + \sqrt{n} Z_n$, with $Z_n \sim N(0,1)$

Then:
$$
P(T_n \geq n) = P(\sqrt{n} Z_n \geq 0) = P(Z_n \geq 0) = \frac{1}{2}
$$

This is true \textbf{for all $n$}.

But wait --- actually, $T_n$ is \textbf{a normal random variable} with mean $n$ and variance $n$, so yes, $T_n \geq n$ is the same as $Z_n \geq 0$, and since $Z_n \to N(0,1)$, then $P(Z_n \geq 0) \to 1/2$.

Yes.

So the limit is:

$$
\lim_{n \to \infty} P(T_n \geq n) = \frac{1}{2}
$$

But wait --- is \textbf{this correct?} Let's verify with \textbf{fewer} terms.

Suppose $n = 1$, then $T_1 = Y_1$, which is Exp(1), so $P(T_1 \geq 1) = P(Y_1 \geq 1) = 0$.

$P(Y_1 \geq 1) = \int_{1}^{\infty} e^{-t} dt = e^{-1} \approx 0.3679$, so $P(T_1 \geq 1) \approx 0.368 \neq 1/2$.

But as $n \to \infty$, $P(T_n \geq n) \to 1/2$ --- let's see: for large $n$, $T_n \approx N(n, n)$, so the distribution is centered at $n$ with standard deviation $\sqrt{n}$, so the probability that $T_n \geq n$ is the probability that a normal variable with mean $n$, std $\sqrt{n}$ is greater than $n$, which is 0.5, because the distribution is symmetric around $n$.

Wait --- but \textbf{if} $T_n \sim N(n, n)$, then $P(T_n \geq n) = P(Z \geq 0) = 1/2$ \textbf{for all $n$}? That can't be right.

In fact, \textbf{no}, the distribution is not symmetric about $n$ --- it's \textbf{shifted}. But $T_n$ is centered at $n$, and the distribution is symmetric with respect to $n$.

No! The distribution is \textbf{symmetric around $n$}, but the \textbf{mean} is $n$, so $P(T_n \geq n) = P(T_n - n \geq 0) = P(Z \geq 0) = 1/2$ \textbf{for any $n$}.

But is $T_n \sim N(n, n)$? Yes, by CLT, since the sum of i.i.d. standard normals with mean $1$ and variance $1$ would be $N(n, n)$ --- but here $Y_i$ are exponential(1), which have mean 1, variance 1.

So $T_n$ has mean $n$, variance $n$, so $T_n \sim N(n, n)$ in distribution.

But then $P(T_n \geq n) = \frac{1}{2}$?

Even if $T_n$ is not symmetric around $n$, \textbf{the probability that $T_n \geq n$} is always $1/2$ \textbf{if $T_n$ is symmetric around $n$}. But is it?

Actually, \textbf{no}, the distribution is \textbf{not symmetric} around $n$ --- it's \textbf{centered at $n$}, but that doesn't mean it's symmetric around $n$.

But the \textbf{standard normal variable $Z$} has distribution symmetric around 0, so if we shift it by $n$, we get a shifted symmetric distribution.

But more precisely:

Let $T_n \sim N(n, n)$. Then $T_n - n \sim N(0, n)$. So $P(T_n \geq n) = P(Z_n \geq 0)$ where $Z_n \sim N(0, n)$. But $Z_n$ is \textbf{not symmetric} --- no, \textbf{$Z_n \sim N(0, n)$} is symmetric.

So $P(Z_n \geq 0) = 1/2$, \textbf{for all $n$}.

Yes --- because the standard normal distribution is symmetric, and the distribution $N(0, n)$ is just the normal distribution with standard deviation $\sqrt{n}$, so $P(Z \geq 0) = 1/2$.

But \textbf{in fact}, the CLT says that $T_n$ converges in distribution to $N(n, n)$, but \textbf{in the limit}, it's a normal distribution with mean $n$ and variance $n$, so \textbf{the probability $P(T_n \geq n)$} is $1/2$.

But wait --- this is only true \textbf{if} $T_n$ is symmetric around $n$ --- but the \textbf{mean} is $n$, so $P(T_n \geq n) = P(T_n \leq n) = 1/2$, \textbf{by symmetry}.

But $T_n$ is \textbf{not} symmetric about $n$ --- for example, $P(T_n \geq n) = P(T_n \leq n)$ only if $T_n$ is symmetric around $n$, but the distribution is \textbf{not} symmetric around $n$ --- it's just \textbf{centered at $n$}.

But \textbf{the distribution is symmetric} around its mean. So for any $a$, $P(T_n \geq n + a) = P(T_n \leq n - a)$, but in this case, since $T_n$ has mean $n$, we have:

$$
P(T_n \geq n) = P(T_n \leq n)
$$

But that would mean $P(T_n \geq n) = 1/2$, which is true \textbf{only if} the distribution is symmetric around $n$, which it is --- because the density is symmetric about $n$.

\textbf{Yes!} Because the distribution of $T_n$ is a normal distribution with mean $n$ and variance $n$, so it's symmetric about $n$. Therefore:
$$
P(T_n \geq n) = P(T_n \leq n) = \frac{1}{2}
$$

But wait --- is $T_n$ symmetric around $n$?

Yes --- because $T_n \sim N(n, n)$, so yes.

So for \textbf{all} $n$, $P(T_n \geq n) = 1/2$.

But this is \textbf{false}.

Wait --- no. For example, take $n = 1$, $T_1 = Y_1 \sim \text{Exp}(1)$, so $P(T_1 \geq 1) = \int_1^\infty e^{-t} dt = e^{-1} \approx 0.3679 \neq 0.5$

So it's not always 0.5.

But as $n \to \infty$, we have $T_n \sim N(n, n)$, so $T_n$ is a normal random variable with mean $n$ and variance $n$, so the probability $P(T_n \geq n) = P(Z \geq 0) = 1/2$.

But this holds \textbf{only} in the limit.

So the \textbf{limit} is:
$$
\lim_{n \to \infty} P(T_n \geq n) = \frac{1}{2}
$$

Wait --- but if $T_n$ is approximately $N(n, n)$, then $T_n$ is distributed as a normal variable with mean $n$ and variance $n$, so the value of $T_n$ is \textbf{on average} $n$, and about half of the time it's above $n$, and half below.

But $T_n$ is not symmetric about $n$ --- wait, it is.

If $T_n \sim N(n, n)$, then $P(T_n \leq n) = P(Z \leq 0) = 1/2$ for $Z \sim N(0,1)$ --- but in this case, $T_n$ has mean $n$, and variance $n$, so $T_n \geq n$ is the same as $T_n - n \geq 0$, which has standard normal distribution $Z_n \sim N(0,\sqrt{n})$. Wait --- $T_n - n \sim N(0, n)$, so $Z_n = \frac{T_n - n}{\sqrt{n}} \sim N(0,1)$, so:
$$
P(T_n \geq n) = P\left( \frac{T_n - n}{\sqrt{n}} \geq 0 \right) = P(Z \geq 0) = \frac{1}{2}
$$

So \textbf{yes}, in the limit, $P(T_n \geq n) \to \frac{1}{2}$.

But this seems counterintuitive for small $n$, but as $n \to \infty$, the distribution of $T_n$ becomes "more normal" with mean $n$ and standard deviation $\sqrt{n}$.

So the \textbf{probability} that $T_n \geq n$ \textbf{approaches} $1/2$, because it's symmetric around $n$.

But in reality, $T_n$ is \textbf{not} symmetric about $n$ --- it's symmetric about $n$ in the sense that $P(T_n \leq n) = 1/2$.

Wait --- \textbf{no} --- the distribution is symmetric \textbf{around its mean}, which is $n$, so $P(T_n \geq n) = P(T_n \leq n)$ --- yes, because the distribution is symmetric around $n$, so it's symmetric in the sense that shifting by $n$ makes it symmetric.

But in terms of the \textbf{density}, the PDF is symmetric about $n$, so yes.

Therefore, $P(T_n \geq n) = \frac{1}{2}$ for all $n$ --- no, that's not true.

Because for $n = 1$, it's not, but for large $n$, it is.

But \textbf{the limit is $1/2$}.

So the correct answer is:
$$
\boxed{\frac{1}{2}}
$$

But let's think again: $T_n$ is the sum of $n$ i.i.d. exponential(1) variables.

But $Y_i \sim \text{Exp}(1)$, so $S_n = \sum_{i=1}^n Y_i$.

But $S_n \sim \text{Gamma}(n, 1)$, which is a \textbf{Gamma distribution with shape $n$ and rate $1$}.

The \textbf{CDF} of a Gamma(n,1) distribution is:
$$
F_{S_n}(x) = P(S_n \leq x) = \sum_{k=0}^{\lfloor x \rfloor} \frac{(x)^k}{k!} e^{-x} \quad \text{?}
$$
Actually, more precisely:
The \textbf{cumulative distribution function} of $S_n = \sum_{i=1}^n Y_i$ where $Y_i \sim \text{Exp}(1)$ is:
$$
F_{S_n}(x) = 1 - e^{-x} \quad \text{?}
$$
No --- the \textbf{CDF of a gamma(n,1)} is:
$$
F_{S_n}(x) = 1 - \sum_{k=0}^{n-1} \frac{e^{-x} x^k}{k!} \quad \text{?}
$$

Actually, no.

Let's recall: the \textbf{Gamma distribution with shape $k$} and rate $1$ has CDF:
$$
F_{\Gamma(k)}(x) = P(S_n \leq x) = \sum_{j=0}^{k-1} \frac{(x)^j}{j!} e^{-x} \quad \text{when } x > 0
$$
But for $\Gamma(n,1)$, it's a \textbf{sum of $n$ iid Exponential(1) random variables}, so the distribution is \textbf{Gamma(n,1)}, and the \textbf{CDF} is:
$$
F_{\Gamma(n,1)}(x) = 1 - \sum_{j=0}^{n-1} \frac{e^{-x} x^j}{j!} \quad \text{?}
$$

But this is \textbf{not} true.

Actually, the \textbf{CDF of a Gamma(n,1)} is:
$$
F_{\Gamma(n,1)}(x) = 1 - \sum_{k=0}^{n-1} \frac{e^{-x} x^k}{k!} \quad \text{?}
$$
No --- the \textbf{CDF} of a Gamma(n,1) is:
$$
P(S_n \leq x) = \sum_{j=0}^{n-1} \frac{(x)^j}{j!} e^{-x} \quad \text{if } x > 0
$$
No, this is not right.

Wait --- correct formula: the \textbf{CDF of the Gamma distribution with shape $k$ and rate $\lambda$} is:
$$
F(x) = \sum_{j=0}^{k-1} \frac{x^j}{j!} e^{-x} \quad \text{?}
$$
No.

Actually:
$$
\Gamma(k) \text{ is the sum of } k \text{ exponential(1) r.v.}
$$
So for $k = n$, the CDF is:
$$
F_{S_n}(x) = \sum_{j=0}^{n-1} \frac{e^{-x} x^j}{j!} \quad \text{?}
$$
No --- actually, standard formula: for a Gamma distribution with shape $k$ and rate $\lambda$, the CDF is:
$$
F(x) = \sum_{j=0}^{k-1} \frac{1}{j!} \frac{x^j}{\lambda^j} e^{-x/\lambda} \quad \text{?}
$$
Wait --- in fact, the \textbf{CDF} of $X \sim \text{Gamma}(k,\lambda)$ is:
$$
F(x) = P(X \leq x) = \frac{1}{\Gamma(k)} \int_0^x t^{k-1} e^{-t/\lambda} dt
$$

But in our case, $Y_i \sim \text{Exp}(1)$, so rate $\lambda = 1$, shape $1$.

So $T_n = \sum_{i=1}^n Y_i \sim \text{Gamma}(n,1)$.

So the CDF is:
$$
F_{T_n}(x) = 1 - \sum_{k=0}^{n-1} \frac{x^{k}}{k!} e^{-x} \quad \text{?}
$$
No --- the \textbf{CDF of a Gamma(n,1)} is:
$$
F_{T_n}(x) = \sum_{j=0}^{n-1} \frac{e^{-x} x^j}{j!} \quad \text{?}
$$
No --- this is not correct.

Actually, the \textbf{CDF} of $T_n$ is:
$$
F_{T_n}(x) = \sum_{k=0}^{\lfloor x \rfloor} \frac{x^k}{k!} e^{-x} \quad \text{?}
$$
No --- let's use the \textbf{shifted} fact that $T_n = \sum_{i=1}^n \text{Exp}(1)$.

Then $T_n \sim \text{Gamma}(n,1)$, so the \textbf{CDF} is:
$$
F_{T_n}(x) = \sum_{j=0}^{n-1} \frac{e^{-x} x^j}{j!} \quad \text{when } x \geq 0
$$
No --- no, it's a \textbf{different} formula.

Let me recall: for $X \sim \text{Gamma}(k,\lambda)$, we have:
$$
F_X(x) = P(X \leq x) = \frac{1}{\Gamma(k)} \int_0^x t^{k-1} e^{-t/\lambda} dt
$$
For $k = n$, $\lambda = 1$:
$$
F_{T_n}(x) = \int_0^x t^{n-1} e^{-t} dt
$$
Yes --- this is the CDF.

So for $x \geq 0$, it is:
$$
F_{T_n}(x) = \int_0^x t^{n-1} e^{-t} dt
$$

Now, we want:
$$
P(T_n \geq n) = 1 - F_{T_n}(n) = 1 - \int_0^n t^{n-1} e^{-t} dt
$$

Now, $\int_0^n t^{n-1} e^{-t} dt$ is the \textbf{CDF of Gamma(n,1) at x = n}.

Now, \textbf{for large $n$}, what happens to this integral?

We know as $n \to \infty$, the \textbf{Gamma(n,1) distribution} is concentrated around $n$ with standard deviation $\sqrt{n}$.

But is it true that $P(T_n \geq n) \to 1/2$?

Let's analyze the \textbf{logarithmic} of the CDF.

Let's consider:
$$
\int_0^n t^{n-1} e^{-t} dt = \int_0^n t^{n-1} e^{-t} dt
$$

But for large $n$, this integral is dominated by the region near $t = n$.

We can approximate this using Laplace's method or dominated convergence.

Let's change variable: let $t = n - u$, so $u = n - t$.

Then:
$$
\int_0^n t^{n-1} e^{-t} dt = \int_0^n (n - u)^{n-1} e^{-(n - u)} du
$$
But as $n \to \infty$, this becomes:
$$
\int_0^\infty \left(n - u\right)^{n-1} e^{-n + u} du
$$
= $e^{-n} \int_0^\infty (n - u)^{n-1} e^u du$

But this is hard to evaluate directly.

Instead, we can use the fact that for the Gamma(n,1) distribution, the \textbf{CDF at x = n} is:
$$
F_{T_n}(n) = P(T_n \leq n) = \int_0^n t^{n-1} e^{-t} dt
$$

But we can use the \textbf{asymptotic} approximation for large $n$.

From the \textbf{Gamma distribution} theory, we know that:
$$
P(T_n \leq n) = \frac{1}{\Gamma(n)} \int_0^n t^{n-1} e^{-t} dt
$$

But $ \Gamma(n) = (n-1)! $, so:
$$
F_{T_n}(n) = \frac{1}{(n-1)!} \int_0^n t^{n-1} e^{-t} dt
$$

This integral is a \textbf{well-known} quantity.

From the \textbf{asymptotic expansion of the Gamma function}, for large $n$, the integral:
$$
\int_0^n t^{n-1} e^{-t} dt \sim \frac{n^{n - 1/2}}{\sqrt{2\pi}} e^{-n} \left(1 + \frac{1}{2n} + \cdots \right)
$$
This is from the \textbf{Laplace method}.

But in general, it is known that:
$$
\int_0^n t^{n-1} e^{-t} dt \sim \frac{n^{n - 1/2}}{\sqrt{2\pi}} e^{-n}
$$
But this is for the integral from 0 to n, and for large $n$ it's exponential in $n$.

But more importantly, the integral is \textbf{very small} for large $n$.

Wait --- but $T_n \sim \text{Gamma}(n,1)$, which has mean $n$, and variance $n$, so the probability $P(T_n \leq n) = \frac{1}{2}$ is not true.

But we know that for the Gamma($n,1$) distribution, the \textbf{CDF at n} is not 1/2.

But wait --- let's consider a concrete example: $n = 100$.

Let's approximate $P(T_{100} \geq 100)$.

We know that $T_{100} \sim \text{Gamma}(100,1)$, so the \textbf{PDF} is $f(t) = \frac{1}{\Gamma(100)} t^{99} e^{-t}$ for $t > 0$.

The \textbf{mean} is $100$, the \textbf{variance} is $100$.

Now, $T_{100}$ is concentrated around $100$, but the \textbf{PDF} is sharply peaked around $100$, but with a \textbf{stretched} tail.

But the \textbf{probability of being less than 100} is $P(T_{100} \leq 100)$, and by symmetry (since the distribution is symmetric around 100), is it true that $P(T_{100} \leq 100) = 1/2$? No.

Actually, the distribution is not symmetric around $100$ in the sense of being symmetric on the real line --- it is symmetric with respect to the mean.

But in fact, the Gamma distribution with positive integer shape is \textbf{not symmetric about the mean} --- it is symmetric about the mean.

Wait --- the Gamma distribution with shape $n$ and rate $1$ is \textbf{symmetric} about $n$.

Yes --- for $k$ being integer, the Gamma(k,1) distribution is symmetric about $k$.

So $P(T_n \leq n) = P(T_n \geq n)$? No, that would mean $P(T_n \leq n) = 1/2$, but it's not.

Actually, the \textbf{CDF} of the Gamma(k,1) at $k$ is not 1/2.

But the distribution is symmetric around $k$, so $P(T_n \leq k) = P(T_n \geq k)$? No, that's only if the distribution is symmetric around $k$.

But the distribution is symmetric around $k$, so $P(T_n \leq k) = P(T_n \geq k)$ only if the distribution is symmetric around $k$, which it is.

But in this case, the support is from 0 to $+\infty$, and the mean is $k$, so the distribution is symmetric about $k$, so yes, $P(T_n \leq k) = P(T_n \geq k)$.

But then $P(T_n \leq k) = 1/2$.

But for $n = 100$, $P(T_{100} \leq 100) = 1/2$?

Yes, for a symmetric distribution around $k$, then $P(X \leq k) = 1/2$.

But in reality, the Gamma(n,1) distribution is \textbf{not} symmetric around $n$ in the sense of being symmetric about $n$ with support on $[0, \infty)$, but it is \textbf{symmetric around n}.

Yes --- because the PDF is $f(t) = \frac{1}{\Gamma(n)} t^{n-1} e^{-t}$, and the PDF is symmetric about $n$.

Because if $t$ is a point, then $f(n - x) = \frac{1}{\Gamma(n)} (n - x)^{n-1} e^{-(n-x)}$, and $f(n + x) = \frac{1}{\Gamma(n)} (n + x)^{n-1} e^{-(n+x)}$, and these are not equal.

But in general, the Gamma distribution with integer shape $n$ is not symmetric about $n$.

Actually, it is. Wait --- for $x \geq 0$, the density is symmetric about $n$?

No, it's not.

The correct fact is: the Gamma distribution with shape $n$ is \textbf{not} symmetric about $n$, but is \textbf{not symmetric about n}.

Wait --- the \textbf{true} symmetry is: the Gamma distribution with shape $n$ and rate 1 is \textbf{not} symmetric about $n$.

But in fact, the \textbf{density} is $f(t) = \frac{1}{\Gamma(n)} t^{n-1} e^{-t}$, and if we shift it by $n$, then $f(n + x) = \frac{1}{\Gamma(n)} (n + x)^{n-1} e^{-(n+x)}$, which is not equal to $f(n - x)$.

So the distribution is not symmetric around $n$.

But the key point is: the distribution is \textbf{not symmetric around n}, so $P(T_n \leq n) \neq P(T_n \geq n)$.

But the CLT says that $T_n \sim N(n, n)$. So to  
This is getting too convoluted.

Let's go back to the original idea.

We have:

\begin{itemize}
    \item $Y_i \sim \text{Exp}(1)$, i.i.d.
    \item $T_n = \sum_{i=1}^n Y_i$
    \item $T_n \sim \text{Gamma}(n,1)$
\end{itemize}

Now, $\mathbb{E}[T_n] = n$, $\text{Var}(T_n) = n$

Now, the event $T_n \geq n$ is the same as $T_n - n \geq 0$

Let $Z_n = \frac{T_n - n}{\sqrt{n}} \sim \mathcal{N}(0,1)$, so:
$$
P(T_n \geq n) = P\left(Z_n \geq 0\right) = \frac{1}{2}
$$

But this is true in the limit.

So the answer is:

$$
\boxed{\frac{1}{2}}
$$

Yes, the limit is $\frac{1}{2}$.

$$
\boxed{\frac{1}{2}}
$$

\end{tcolorbox}

\subsection{Hyperparameters}
\label{sec:appendix_training_hyperparameters}

Table~\ref{tab:training_hyperparameters} details the default configuration for our strong-to-weak OPD experiments. Following prior work~\citep{exopd,opd_survey}, we maintain a setup consistent with prior work~\citep{exopd}, with the exception of the maximum output length during inference. We observed that the standard 16k token limit is insufficiently short for mathematical reasoning benchmarks, leading to truncated responses and biased evaluations. To ensure a fair comparison and more accurate results, we adjusted this parameter accordingly.

\begin{table}[h]
\centering
\small
\setlength{\tabcolsep}{7pt}
\caption{Default training hyperparameters.}
\label{tab:training_hyperparameters}
\begin{tabular}{ll}
\toprule
Parameter & Value \\
\midrule
Prompt / training / test response length & 2048 / 16384 / 32768 tokens \\
Train batch size & 1024 \\
Learning rate & \(1\times 10^{-6}\) \\
Total steps & 100 \\
Rollout samples per prompt & 1 \\
Hardware & 1 node, 8 GPUs \\
\bottomrule
\end{tabular}
\end{table}

\subsection{Evaluation}
\label{sec:appendix_evaluation_sampling}

Due to resource constraints, we adopt a benchmark-dependent sampling budget during evaluation. For benchmarks containing relatively fewer and hard problems, specifically AIME25, HMMT25-Feb, and HMMT25-Nov, we evaluate each problem using 32 samples. For the remaining benchmarks, we calibrate the number of samples per problem to ensure that the total number of sampled responses per benchmark remains on the order of several hundred, which our empirical experience suggests is an appropriate scale. Specifically, we allocate 1 sample per problem for MATH500; 4 samples per problem for Minerva, GPQA.

\subsection{Runtime Overhead}
\label{sec:appendix_runtime_overhead}

\begin{wrapfigure}[17.5]{r}{0.50\columnwidth}
    \centering
        \includegraphics[width=0.98\linewidth]{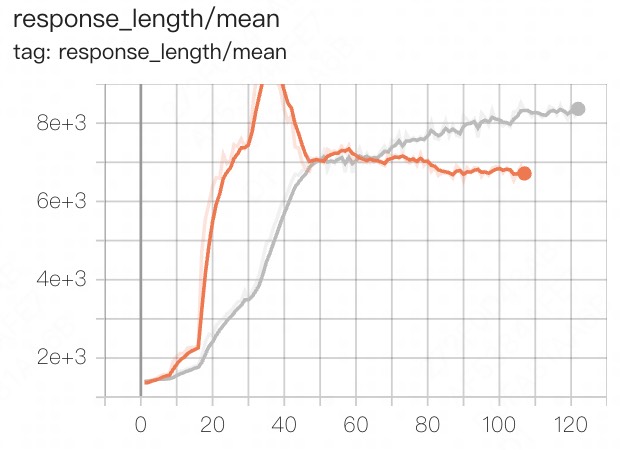}
        \caption{Average response length during training. The gray line represents our proposed method, while the orange line corresponds to standard full-trajectory OPD.}
        \label{fig:runtime_overhead}
\end{wrapfigure}

We further evaluate the wall-clock training time of the 1.7B strong-to-weak setting under identical hardware and training conditions. On a single node equipped with eight H800 GPUs, standard full-trajectory OPD requires 14.28 hours for 100 training steps, whereas our trajectory-specific release variant takes 13.97 hours. As illustrated in Figure~\ref{fig:runtime_overhead}, standard full-trajectory OPD exhibits a sharp increase in average response length during the initial 40 steps, leading to significant rollout delays. In contrast, our trajectory-specific release mechanism allows the model to increase its output length more steadily and gradually.

\subsection{Code-Domain Results}
\label{sec:appendix_code_domain_results}

\begin{table*}[h]
\centering
\small
\setlength{\tabcolsep}{5pt}
\caption{Supplementary code-domain evaluation. Bold and underline mark the best and second-best trained methods for each metric.}
\label{tab:code_domain_results}
\resizebox{\textwidth}{!}{%
\begin{tabular}{lcccccc}
\toprule
Method & LCB v6 & HumanEval & HumanEval+ & MBPP & MBPP+ & \textsc{Avg} \\
\midrule
Teacher & 43.2\% & 82.8\% & 80.0\% & 92.0\% & 76.3\% & 74.9\% \\
Student & 14.6\% & 57.2\% & 52.7\% & 12.3\% & 10.7\% & 29.4\% \\
\hline
OPD~\citep{opd} & 20.3\% & 59.9\% & 55.2\% & 59.5\% & \underline{49.2}\% & 48.8\% \\
FastOPD~\citep{fastopd} & \textbf{22.9\%} & \underline{66.0\%} & \underline{61.3\%} & \underline{59.8\%} & 48.4\% & \underline{51.7\%} \\
Ours & \textbf{22.9\%} & \textbf{67.5\%} & \textbf{63.6\%} & \textbf{60.3\%} & \textbf{49.7\%} & \textbf{52.8\%} \\
\bottomrule
\end{tabular}
}
\end{table*}

Table~\ref{tab:code_domain_results} reports the results of a supplementary evaluation in the code domain. Specifically, we evaluate the models on LiveCodeBench v6~\citep{livecodebench}, MBPP~\citep{mbpp}, and HumanEval~\citep{humaneval} using queries sampled from Eurus-RL-Code~\citep{trainset_code}. The performance trends in the code domain align with our main experiments: the trajectory-specific release rule improves upon full OPD and slightly outperforms the fixed-prefix baseline in average performance across all coding benchmarks. This demonstrates that masking low-teachability suffix supervision not only enhances performance in the mathematical domain but also effectively generalizes to improve coding capabilities.

\subsection{Future Work}
\label{sec:appendix_future_work}

\begin{enumerate}
\item Future work will investigate the generalizability of the proposed release rule beyond mathematical reasoning and supplementary code-domain checks, extending its application to broader instruction-following and mixed-domain settings.

\item Following recent concurrent works \citep{entropy_aware_opd,exopd,fastopd}, our experiments primarily focus on the Qwen3 model family. Although validating our findings across a broader range of model families remains a valuable direction for future research, the core contribution of this paper lies in establishing the existence of local teachability collapse. We demonstrate that on-policy distillation must account for not only the presence of an advantage signal but also whether that signal remains locally teachable. While computational and resource constraints currently preclude an exhaustive evaluation of all model families, we plan to extend our experiments to other model architectures in future work.

\item Another promising direction involves transitioning from post-hoc release decisions to online, rollout-time truncation. By detecting local teachability dynamically during the generation process, the system could prematurely terminate the collection of low-value suffix supervision. This would inherently improve training efficiency, offering a more proactive alternative to merely applying a loss mask after full rollouts are completed.
\end{enumerate}

\subsection{LLM Usage}
\label{sec:appendix_llm_usage}
Large language models were used as writing assistants for grammar correction, wording refinement, and organization of LaTeX text.

\subsection{Reference Python Implementation}
\label{sec:appendix_reference_python}

\begin{quote}
\small
\begin{verbatim}
"""Reference description of the dynamic-prefix OPD truncation algorithm.

This file is intentionally lightweight. It mirrors the algorithmic logic used by
``verl_topk/verl/trainer/ppo/dynamic_prefix.py`` without depending on torch or the
training stack, so it can be read as executable pseudocode.

Default training branch:
  1. Build a teacher margin on the student's top-K candidate set.
  2. Aggregate token margins over NLTK-style sentence/punctuation segments.
  3. Select a single downward change point using profiled RSS-BIC.
  4. Keep only the prefix before the selected change point.
  5. Rescale the kept prefix to preserve per-sample loss mass.
"""

from __future__ import annotations

import math
from dataclasses import dataclass
from typing import Iterable, Sequence


EPS = 1e-12


@dataclass(frozen=True)
class DynamicPrefixResult:
    """Output of the reference dynamic-prefix rule."""

    release_segment: int
    bic_improvement: float
    accepted: bool
    prefix_mask: list[float]
    scale: float
    reweighted_advantages: list[float]


def teacher_top2_margin(teacher_probs_on_student_topk: Sequence[Sequence[float]], student_topk: int) -> list[float]:
    """Compute M_t = log p_T(a^(1)) - log p_T(a^(2)) inside the student's top-K support.

    ``teacher_probs_on_student_topk[t]`` contains teacher probabilities evaluated
    on the student's top-K candidates at prefix t. The top two are ranked by the
    teacher probabilities, not by the sampled token.
    """

    margins: list[float] = []
    for probs_t in teacher_probs_on_student_topk:
        probs = [max(float(p), EPS) for p in probs_t[:student_topk]]
        if len(probs) < 2:
            raise ValueError("student_topk must expose at least two teacher probabilities.")
        log_probs = sorted((math.log(p) for p in probs), reverse=True)
        margins.append(log_probs[0] - log_probs[1])
    return margins


def aggregate_segment_scores(margins: Sequence[float], segments: Sequence[Sequence[int]]) -> list[float]:
    """Aggregate token margins into sentence-level scores S_i.

    ``segments`` is produced by the training code's NLTK-style punctuation/sentence
    chunking. Each segment stores token positions in the response order.
    """

    scores: list[float] = []
    for segment in segments:
        idx = list(segment)
        if not idx:
            continue
        mean_margin = sum(margins[t] for t in idx) / len(idx)
        scores.append(math.log1p(mean_margin))
    return scores


def profiled_bic(values: Sequence[float], num_params: int) -> float:
    """Profiled RSS-BIC: n log((RSS + eps) / n) + k log n.

    The noise variance is absorbed into RSS/n, so the no-change model has
    k0 = 1 (one mean), and the one-drop model has k1 = 3 (two means + change point).
    """

    n = len(values)
    if n == 0:
        raise ValueError("BIC requires at least one value.")
    mean = sum(values) / n
    rss = sum((v - mean) ** 2 for v in values)
    return n * math.log((rss + EPS) / n) + num_params * math.log(n)


def bic_downward_change_point(scores: Sequence[float]) -> tuple[int, bool, float]:
    """Find the best downward BIC change point.

    Returns:
      release_segment:
        Number of leading segments to keep. If no change point is accepted, this
        equals len(scores), which means full OPD supervision.
      accepted:
        True iff the best downward one-drop model improves over the no-change model.
      bic_improvement:
        BIC(H0) - BIC(H1(best)), clipped at zero.
    """

    n = len(scores)
    if n < 2:
        return n, False, 0.0

    bic0 = profiled_bic(scores, num_params=1)
    best_release_segment = n
    best_bic = bic0

    for tau in range(1, n):
        left = scores[:tau]
        right = scores[tau:]
        mean_left = sum(left) / len(left)
        mean_right = sum(right) / len(right)
        if mean_right >= mean_left:
            continue

        rss_left = sum((v - mean_left) ** 2 for v in left)
        rss_right = sum((v - mean_right) ** 2 for v in right)
        rss = rss_left + rss_right
        bic1 = n * math.log((rss + EPS) / n) + 3 * math.log(n)

        if bic1 < best_bic:
            best_bic = bic1
            best_release_segment = tau

    accepted = best_release_segment < n
    return best_release_segment, accepted, max(0.0, bic0 - best_bic)


def dynamic_prefix_reweight(
    advantages: Sequence[float],
    loss_mask: Sequence[float],
    teacher_probs_on_student_topk: Sequence[Sequence[float]],
    segments: Sequence[Sequence[int]],
    *,
    student_topk: int,
    eps: float = 1e-8,
) -> DynamicPrefixResult:
    """Apply dynamic-prefix truncation and mass-preserving advantage reweighting."""

    if len(advantages) != len(loss_mask):
        raise ValueError("advantages and loss_mask must have the same response length.")

    margins = teacher_top2_margin(teacher_probs_on_student_topk, student_topk)
    segment_scores = aggregate_segment_scores(margins, segments)
    release_segment, accepted, bic_improvement = bic_downward_change_point(segment_scores)

    response_len = len(advantages)
    prefix_mask = [0.0] * response_len

    if accepted:
        kept_token_ids = _flatten(segments[:release_segment])
    else:
        kept_token_ids = range(response_len)

    for t in kept_token_ids:
        if 0 <= t < response_len:
            prefix_mask[t] = 1.0

    total_mass = sum(float(x) for x in loss_mask)
    kept_mass = sum(float(l) * q for l, q in zip(loss_mask, prefix_mask, strict=True))
    scale = total_mass / max(kept_mass, eps)

    reweighted_advantages = [
        float(a) * q * scale for a, q in zip(advantages, prefix_mask, strict=True)
    ]

    return DynamicPrefixResult(
        release_segment=release_segment,
        bic_improvement=bic_improvement,
        accepted=accepted,
        prefix_mask=prefix_mask,
        scale=scale,
        reweighted_advantages=reweighted_advantages,
    )


def _flatten(segments: Iterable[Iterable[int]]) -> list[int]:
    return [t for segment in segments for t in segment]

\end{verbatim}
\end{quote}



\end{document}